\documentclass{article}
\providecommand{\papermode}{preprint}

\usepackage{iclr2027_conference,times}
\usepackage{etoolbox}
\ifdefstring{\papermode}{submission}{%
  \iclrfinalfalse
}{%
  \ifdefstring{\papermode}{preprint}{%
    \iclrfinalcopy
    \AddToHook{cmd/@maketitle/after}{\lhead{Preprint}}
  }{%
    \ifdefstring{\papermode}{accept}{%
      \iclrfinalcopy
    }{%
      \PackageError{paper-mode}{Unknown paper mode: \papermode}%
        {Choose submission, preprint, or accept at the top of main.tex.}
    }%
  }%
}
\usepackage{amsmath,amssymb,amsthm,graphicx,booktabs,microtype,array}
\usepackage{hyperref,url,xurl,placeins,xcolor,wrapfig}
\usepackage[capitalise,noabbrev,nameinlink]{cleveref}
\crefname{figure}{Figure}{Figures}
\Crefname{figure}{Figure}{Figures}
\crefname{table}{Table}{Tables}
\Crefname{table}{Table}{Tables}
\crefname{section}{Section}{Sections}
\Crefname{section}{Section}{Sections}
\crefname{subsection}{Section}{Sections}
\Crefname{subsection}{Section}{Sections}
\crefname{subsubsection}{Section}{Sections}
\Crefname{subsubsection}{Section}{Sections}
\crefname{appendix}{Appendix}{Appendices}
\Crefname{appendix}{Appendix}{Appendices}
\crefname{subappendix}{Appendix}{Appendices}
\Crefname{subappendix}{Appendix}{Appendices}
\crefname{subsubappendix}{Appendix}{Appendices}
\Crefname{subsubappendix}{Appendix}{Appendices}
\crefname{equation}{Equation}{Equations}
\Crefname{equation}{Equation}{Equations}

\hypersetup{hidelinks,pdfauthor={},pdftitle={Thinking in Depth: Retrospective Inference for Tabular Foundation Models}}
\newcommand{\Retro}{\textsc{Retro}}
\newcommand{\E}{\mathbb{E}}
\newcommand{\R}{\mathbb{R}}

\newcommand{\rms}{\operatorname{RMSNorm}}

\crefname{proposition}{Proposition}{Propositions}
\Crefname{proposition}{Proposition}{Propositions}
\title{Thinking in Depth: Retrospective Inference for Tabular Foundation Models
}
\author{%
\textbf{Hao-Run Cai, Si-Yang Liu, Zi-Jian Cheng, Kun-Yang Yu, Jin-Hao Sheng, Guo Yu,} \\
\textbf{Chonghan Liu, Zhi Zhou, Jun-Peng Jiang, Lan-Zhe Guo, Han-Jia Ye} \\
{\normalfont School of Artificial Intelligence, Nanjing University}\\
{\normalfont National Key Laboratory for Novel Software Technology, Nanjing University}\\
}
\ifdefstring{\papermode}{submission}{}{%
  \hypersetup{pdfauthor={Hao-Run Cai, Si-Yang Liu, Zi-Jian Cheng, Kun-Yang Yu, Jin-Hao Sheng, Guo Yu, Chonghan Liu, Zhi Zhou, Jun-Peng Jiang, Lan-Zhe Guo, Han-Jia Ye}}%
}
\begin{document}
\maketitle
\begin{abstract}
Tabular foundation models (TFMs) are pretrained across diverse tabular tasks and make predictions on a new table at inference time using its labeled examples as context. Most recent TFMs perform such in-context prediction with stacked Transformer layers, repeatedly transforming how examples are represented and compared. By tracing individual queries through several strong TFMs, we find that predictive refinement is highly uneven across depth and is often concentrated in later layers. This uneven refinement motivates us to reconsider how intermediate representations are constructed and reused throughout the network.
We introduce \Retro{}, a tabular foundation model based on \emph{retrospective inference}, where later stages can explicitly revisit and recombine intermediate information produced earlier in the network. \Retro{} organizes this process around two complementary operations: \emph{which intermediate information to revisit, and how the resulting contextual update should be shaped for each query}. Attention Residuals address the former by adaptively reweighting contributions from different depths, while query-conditioned Gated Attention addresses the latter by modulating the attention output element-wise across representation dimensions.
Our analysis shows that \Retro{} shifts predictive refinement earlier and more broadly across depth, with different stages revising different subsets of queries in a pattern suggestive of multi-view refinement.
Across TabArena, TALENT, and RelArena, \Retro{} ranks among the top three and lies on the Pareto frontier. These results indicate that directly reusing intermediate representations provides a practical way to better exploit depth in TFMs.\looseness=-1
\end{abstract}

\section{Introduction}

\begin{wrapfigure}{r}{0.48\textwidth}
  \centering
  \vspace{-34.5pt}
  \makebox[\linewidth]{\includegraphics[width=\linewidth]{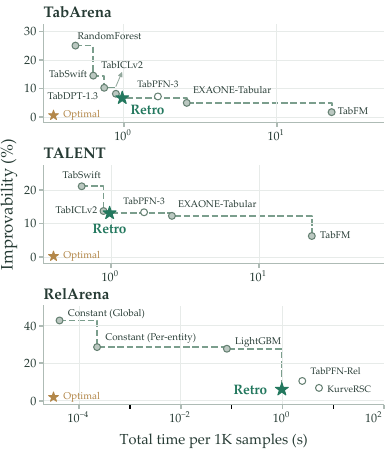}}
  \vspace{-19.8pt}
  \caption{\Retro{} lies on the Pareto frontier of three mainstream benchmarks.}
  \label{fig:benchmark-pareto}
  \vspace{-20pt}
\end{wrapfigure}

Tabular foundation models (TFMs) are reshaping tabular learning from training a separate predictor for each dataset toward reusing a pretrained prediction procedure across heterogeneous tables.
By pretraining over diverse tasks, TFMs can adapt to a new table at inference time using its labeled examples as context, without requiring a task-specific model to be learned from scratch~\citep{hollmann2023tabpfn,hollmann2025tabpfn,qu2025tabicl}.
Recent models have demonstrated increasingly strong performance across a broad range of tabular tasks~\citep{qu2026tabiclv2,grinsztajn2026tabpfn3,eo2026exaone}, establishing TFMs as a promising general-purpose paradigm for tabular prediction.

\begin{figure}[t]
  \centering
  \vspace{-0pt}
  \includegraphics[width=\linewidth]{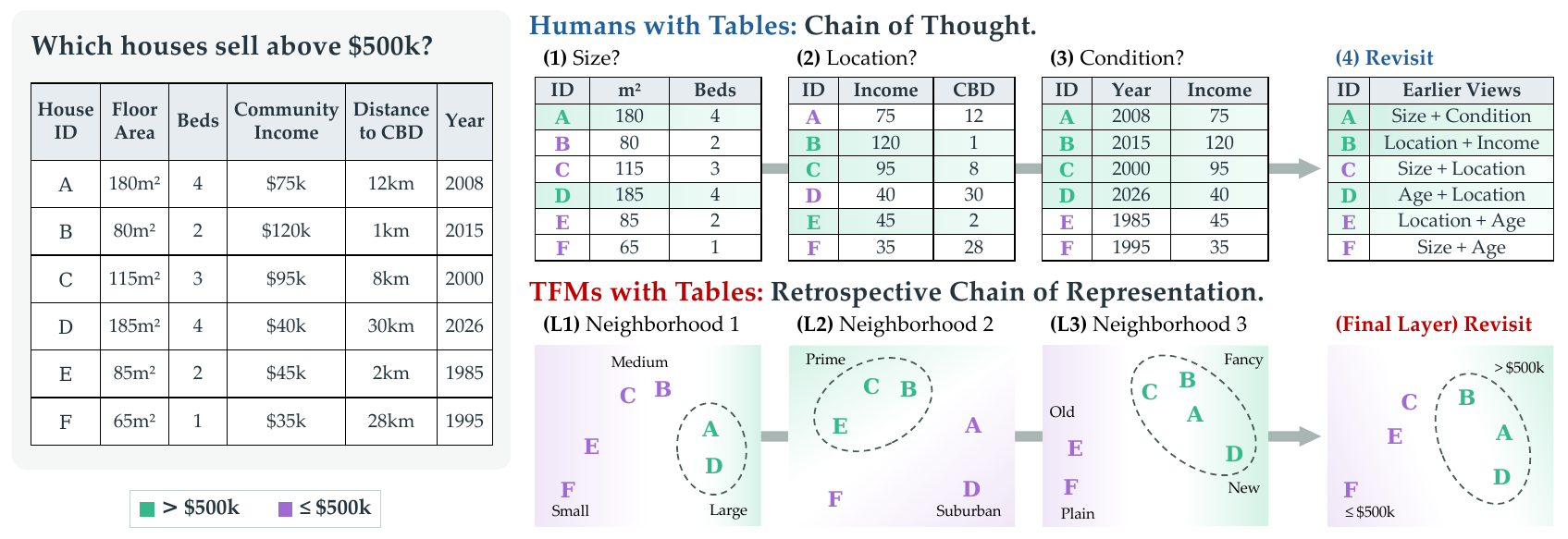}
  \vspace{-10pt}
  \caption{\textbf{A housing example of complementary neighborhood views and retrospective integration.} \textbf{Top:} a person considers size, location, and condition, then revisits these assessments to reach a decision. \textbf{Bottom:} the TFM schematic places the same homes in different neighborhoods across representations and integrates earlier information in the final representation. This illustrates a possible organization of tabular inference.}
  \label{fig:intro}
\end{figure}

In most recent in-context TFMs, labeled support examples and unlabeled queries pass through a stack of Transformer layers before producing the final prediction.
Several works have examined what happens along this layered computation.
\citet{ye2025closer} show that representations formed within a TFM contain task-conditioned predictive structure and can support downstream prediction, while \citet{balef2026onelayer} reveal substantial redundancy and non-uniform contributions across layers through probing and layer interventions.
These studies begin to characterize how predictions emerge through the internal computation of TFMs rather than only at their final output.

We take a complementary, sample-level view by tracing how individual queries change as they pass through several strong TFMs.
We find that predictive refinement is highly uneven across depth and, in some models, is strongly concentrated in later layers.
Different queries are also revised at different stages, revealing a much less uniform process than final aggregate performance alone would suggest.
These observations suggest that the issue may not simply be what each layer computes, but also how information produced along the way remains available to later prediction.
This motivates a different organization of intermediate computation, in which earlier representations can be reconsidered rather than influencing subsequent stages only through the latest accumulated state.

We call this organization \emph{retrospective inference}.
Instead of viewing prediction as a sequence in which each representation simply gives way to the next, retrospective inference keeps intermediate information available as the prediction evolves.
The goal is to let later computation selectively draw on representations formed at different stages, so that useful information can be retained and recombined as different queries are progressively refined.

This perspective leads to two coupled design requirements.
First, when several intermediate representations remain available, the model must determine which of them are relevant to the current sample.
Second, after contextual information is computed from these representations, the model must determine how that update should modify the current query representation.
Accordingly, we design \Retro{} with a retrospective aggregation path that adaptively reweights contributions from different depths, and a query-conditioned gating mechanism that modulates newly computed contextual information across representation dimensions.
We instantiate the former with Attention Residuals~\citep{kimi2026attnres} and the latter with Gated Attention~\citep{qiu2025gated}.
Together, they allow the model to adapt both the historical information it reuses and the way new contextual information is incorporated for each query. \looseness=-1

Our analysis shows that \Retro{} shifts predictive refinement earlier and distributes it more broadly across depth, with different stages revising different subsets of queries in a pattern suggestive of multi-view refinement.
Across TabArena~\citep{erickson2025tabarena}, TALENT~\citep{ye2024closer}, and RelArena~\citep{RelArena}, \Retro{} ranks in the top three among the evaluated methods, while lying on the estimated performance--cost Pareto frontier of each benchmark.
These results highlight the value of treating intermediate representations as reusable resources for subsequent inference. Our contributions are threefold:
\setlength{\leftmargini}{1.2em}
\setlength{\labelsep}{0.45em}
\begin{itemize}
\setlength{\itemsep}{2pt}
\setlength{\parskip}{0pt}
  \item We characterize sample-level predictive dynamics in strong TFMs, complementing prior representation- and layer-level analyses, and show that refinement is highly uneven across depth and often concentrated in later stages.
  \item We propose \emph{retrospective inference} and instantiate it in \Retro{} through adaptive reuse of intermediate representations and query-conditioned modulation of contextual updates, leading to earlier and more distributed prediction refinement across stages.
  \item We evaluate \Retro{} on three benchmarks spanning tabular classification, regression, and relational prediction, obtaining top-three overall Elo rankings among the evaluated methods.
\end{itemize}

\section{Related Work}

\noindent{\bf Tabular machine learning.}
Tabular supervised learning predicts a target for each row from a set of heterogeneous feature columns, typically including numerical and categorical variables.
Most dataset-specific tabular predictors fall broadly into two families.
Tree-based methods, particularly gradient-boosted decision trees, include XGBoost~\citep{chen2016xgboost}, LightGBM~\citep{ke2017lightgbm}, and CatBoost~\citep{prokhorenkova2018catboost}.
Deep tabular methods learn task-specific predictors with neural architectures, including FT-Transformer~\citep{gorishniy2021revisiting}, RealMLP~\citep{holzmuller2024realmlp}, ModernNCA~\citep{ye2025MNCA}, TabM~\citep{Yury2024TabM}, and TabPack~\citep{gorishniy2026tabpack}.
The relative strengths of these methods have been systematically examined in large-scale empirical studies and benchmarks~\citep{grinsztajn2022tree,mcelFresh2023tabzilla,liu2025talent,erickson2025tabarena}.
Broader surveys further organize the architectures, learning paradigms, and generalization regimes of tabular learning methods~\citep{borisov2024survey, Jiang2026Survey}.

\noindent{\bf Tabular foundation models.}
Tabular foundation models (TFMs) move beyond dataset-specific fitting by pretraining a prediction procedure across many tasks and applying it to unseen tables using labeled examples as context~\citep{BreugelS24Position}.
TabPFN~\citep{hollmann2023tabpfn} established this paradigm with a Transformer that performs in-context prediction over labeled and query examples.
Subsequent TFMs mainly differ in how they represent a table and exchange information across features and samples.
One family retains fine-grained feature or cell states while exchanging information across feature and sample axes, including TabPFN v2~\citep{hollmann2025tabpfn}, LimiX~\citep{zhang2025limix}, and EXAONE-Tabular~\citep{eo2026exaone}.
Another first aggregates feature information into row representations and then performs row-level in-context learning, including TabICL~\citep{qu2025tabicl}, TabICLv2~\citep{qu2026tabiclv2}, TabPFN-3~\citep{grinsztajn2026tabpfn3}, and TabFM~\citep{kong2026tabfm}.
Efficient direct row-wise modeling provides another design, as exemplified by TabSwift~\citep{liu2026tabswift}.
Orthogonal to architecture, richer synthetic priors and downstream adaptation improve transfer across heterogeneous tasks, as explored by TabForestPFN~\citep{TabForestPFN}, Mitra~\citep{zhang2025mitra} and Mitra-v2~\citep{tao2026mitrav2}.
\Retro{} follows the feature-to-row, row-level ICL hierarchy, but redesigns representation flow within the ICL stack so that intermediate contributions remain directly reusable by later stages.

Recent work has also begun to examine how TFMs arrive at their predictions beyond final predictive accuracy.
\citet{ye2025closer} analyze task-conditioned representations in TabPFN v2 and study the predictive utility and reuse of intermediate representations.
\citet{balef2026onelayer} investigate layer-wise inference dynamics across multiple TFMs using probing and layer interventions, revealing substantial redundancy across depth.
\citet{bilos2026mechanistic} use causal interventions to study similarity-based in-context prediction mechanisms across contemporary TFMs.
In our work, such analysis serves as a design motivation rather than the endpoint: we trace sample-level prediction changes, observe uneven refinement across depth, and use this phenomenon to motivate retrospective reuse of intermediate computation.\looseness=-1

\noindent{\bf Residual connections.}
The study of inference across depth also raises questions about how intermediate information is passed between layers. Standard residual connections~\citep{he2016deep} facilitate optimization through additive updates, allowing earlier computations to contribute to later states.
Subsequent designs introduce learned gates \citep{srivastava2015highway} or residual scaling \citep{bachlechner2021rezero} to control the propagation of information and improve training.
Hyper-Connections~\citep{zhu2024hyperconnections} extend this design to multiple interacting residual streams, while Manifold-Constrained Hyper-Connections (mHC)~\citep{DBLP:journals/corr/abs-2512-24880} constrain their mixing matrices to stabilize signal propagation across depth.
A complementary direction provides explicit access to earlier computations. DenseNet~\citep{huang2017densely} concatenates earlier feature maps, while Attention Residuals \citep{kimi2026attnres} use input-dependent weights to aggregate contributions from different depths. Delta Attention Residuals \citep{luo2026deltaattnres} further investigate the representations used as sources for this aggregation.
In tabular modeling, Xiaomi-TabLDM~\citep{wang2026xiaomi} incorporates lightweight Attention Residuals into its architecture. \Retro{} builds on block Attention Residuals to organize information flow within the ICL stack, and examines how retrospective access relates to sample-level prediction changes across depth.

\section{Preliminaries}
\label{sec:preliminary}

\noindent{\bf Tabular in-context learning.}
Let $S=\{(\mathbf{x}_j,y_j)\}_{j=1}^{n_s}$ be a labeled support set and $Q=\{\mathbf{x}_i^q\}_{i=1}^{n_q}$ a set of unlabeled queries from the same table. The target space is $\mathcal{Y}=\{1,\ldots,K\}$ for classification and $\mathcal{Y}=\R$ for regression. A tabular foundation model (TFM) is pretrained across tasks with different feature schemas and targets. At inference time, it predicts $p_\theta(y \mid \mathbf{x}_i^q,S)$ using observed support labels while keeping its pretrained parameters $\theta$ fixed; query labels are withheld. In architectures such as TabICLv2 \citep{qu2026tabiclv2}, this process consists of two stages: constructing a representation for each row, then performing in-context prediction over these row representations.

\noindent{\bf Row representations.}
Following TabICLv2, we use column-wise embedding and row-wise feature aggregation to construct the inputs to the ICL stack. The column encoder produces context-dependent feature representations, and the row encoder aggregates them into a fixed-width vector $h_{0i}\in\R^d$ for each sample. For $n=n_s+n_q$ support and query rows, these vectors form $H_0\in\R^{n\times d}$. Observed support labels are incorporated into the representations used for ICL, enabling the subsequent layers to condition predictions on the current task.

\noindent{\bf Multi-layer in-context prediction.}
An $L$-block Transformer processes $H_0$ using the labeled support set as context. Let $H_\ell\in\R^{n\times d}$ collect the row representations after block $\ell$, with row $i$ denoted by $h_{\ell i}$. Each block applies attention followed by a feed-forward network, both with residual connections.
Before attention, layer normalization produces $Z_\ell=\operatorname{LN}(H_{\ell-1})$, where $\operatorname{LN}$ denotes layer normalization applied independently to each row. Let $Z_\ell^S$ contain the normalized support rows. For clarity, we write the attention computation for a single head and omit layer indices on its parameters:
\begin{equation}
  \begin{aligned}
  O_\ell &= \operatorname{softmax}\!\left(
    \frac{(Z_\ell W_Q^\top)(Z_\ell^S W_K^\top)^\top}{\sqrt{d_k}}
    \right)Z_\ell^S W_V^\top,\\
  A_\ell &= O_\ell W_O^\top + \mathbf{b}_O^\top.
  \end{aligned}
  \label{eq:icl-attention}
\end{equation}
Here $W_Q$, $W_K$, and $W_V$ are learned query, key, and value projections; $d_k$ is the query/key dimension. The softmax normalizes each receiving row's scores over the support rows. $W_O$ and $\mathbf{b}_O$ are the output projection and its bias, shared across rows. In multi-head attention, $O_\ell$ concatenates the outputs of all heads before this projection. The update $A_\ell$ is added to $H_{\ell-1}$, followed by a pre-normalized feed-forward residual update. After $L$ blocks, a prediction head maps the final query representations to outputs. \looseness=-1

\begin{figure}[!t]
  \centering
  \includegraphics[width=\linewidth]{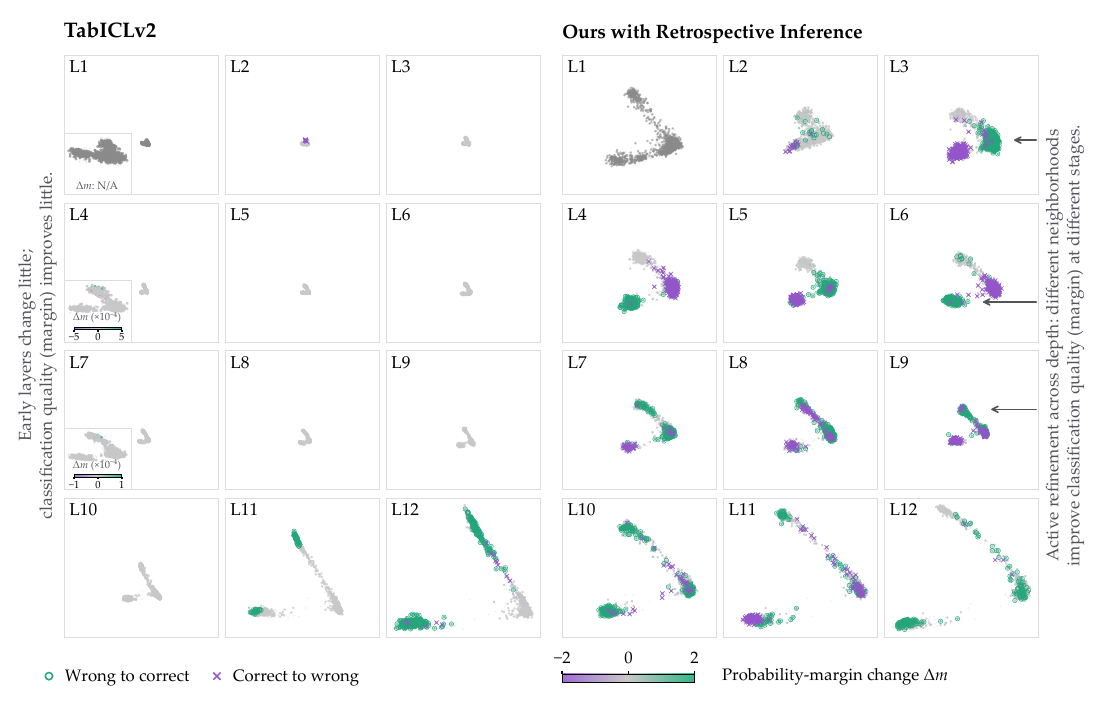}
  \vspace{-20pt}
  \caption{\textbf{Layer-wise changes in predictive margin for TabICLv2 and a retrospective variant on SDSS17.} The left and right grids show query representations across all 12 layers of TabICLv2 and our retrospective variant, respectively. Within each model, representations are projected onto a fixed PCA basis fitted to its final-layer support embeddings; coordinates are therefore comparable across layers within a model, but not between models. Color indicates the change in true-class probability margin from the preceding layer ($\Delta m$), with green denoting an increase and purple a decrease. Circles mark queries changing from incorrect to correct, and crosses mark the reverse. L1 is shown in gray because no preceding layer is measured. Insets enlarge L1, L4, and L7; the L4 and L7 insets use separately labeled, narrower color scales to reveal small early-layer changes. TabICLv2 exhibits limited directly readable changes through much of its depth, whereas the retrospective variant shows changes for different queries across more stages. Full four-model visualizations and aggregate results over 38 datasets appear in \cref{app:baseline-views,app:depth-dynamics}.}
  \label{fig:representation-comparison}
\end{figure}

\section{How Do TFMs Refine Predictions Across Depth?}
\label{sec:dynamics}

Following \citet{ye2025closer}, we extract support embeddings in a held-out query role and fit a linear readout on their final-layer representations. Applying the same readout to $h_{\ell i}$ across depth lets us track each query's predictive state and true-class probability margin. We analyze TabICLv2, TabPFN-3, and EXAONE-Tabular on the first official split of 38 TabArena classification datasets \citep{erickson2025tabarena}. Extraction, margin definitions, and visualization details are given in \cref{app:protocol}.

We find that predictive refinement is often concentrated in a small portion of the network. In \cref{fig:representation-comparison} (left), TabICLv2 shows little visible predictive change through much of its depth, followed by a pronounced late transition. 
Across datasets, the final third accounts for 84.4\% of its absolute margin movement, compared with 51.5\% for TabPFN-3 and 61.1\% for EXAONE-Tabular. Across all three models, predictive refinement is unevenly distributed across depth, although its concentration varies across architectures.

These observations motivate organizing depth around more differentiated contributions to prediction. Successive representations could emphasize different sample relationships, while keeping their results available for later integration. 
This diagnostic tracks how predictive information becomes accessible to a common final-layer readout across depth. Our design question is how to make intermediate results explicitly reusable as the focus of subsequent computation changes.

\section{Retrospective Inference for Tabular Foundation Models}
\label{sec:retrospection}

The findings in \cref{sec:dynamics} raise a question about representation
learning: how does the way intermediate information is reused affect
the contributions learned across depth? We introduce \emph{retrospective
inference}, which makes earlier contributions directly available to later
computation. Attention Residuals implement this reuse, while Gated Attention
modulates the contextual updates produced within the resulting stack.

\subsection{Retrospective Reuse with Attention Residuals}

In a standard residual stack, each sublayer reads a state that accumulates
preceding updates. An early contribution can influence subsequent prediction,
but only as part of this evolving state; it is no longer available as a
separate source that later layers can selectively weight. Consequently,
its predictive value depends on how subsequent transformations retain
and use it. This organization permits layer-wise specialization, but does
not explicitly support selecting an earlier contribution independently
of those accumulated after it.

Retrospective inference changes this relationship between intermediate
computation and prediction. By retaining earlier contributions as separate
sources, it allows a stage to provide information that remains useful
when selected later, even if that information is insufficient for prediction
on its own. We hypothesize that this reuse pathway favors complementary
contributions across depth: different stages can refine different sample
relationships without requiring their results to remain separately
recoverable from a single accumulated state. For tabular ICL, this means
that a representation useful for comparing some queries with their support
examples can remain accessible while later stages refine other comparisons.

We realize this access through block Attention Residuals
\citep{kimi2026attnres}. The source bank contains the initial row encoding,
sums of residual updates from completed two-layer groups, and the current
group's partial sum. At reading site $t$, let $\mathcal B_t$ denote the
available sources and $b_i$ the representation of row $i$ in source $b$.
A learned vector $w_t$ produces row-dependent weights:
\begin{equation}
  \alpha_{tbi}=
  \frac{\exp(\langle w_t,\rms_t(b_i)\rangle)}
       {\sum_{b'\in\mathcal B_t}\exp(\langle w_t,\rms_t(b'_i)\rangle)},
  \qquad
  \widetilde h_{ti}=\sum_{b\in\mathcal B_t}\alpha_{tbi}b_i.
  \label{eq:attnres}
\end{equation}
The scorer is shared across rows but learned separately at each reading
site. Weights are recomputed from each row's candidate contributions.
RMS normalization applies only to scoring; the values remain unnormalized.
Independent readers precede every feed-forward sublayer and every attention
sublayer after the first. Sublayer outputs accumulate within the current
group, and completed groups remain available to later readers.
The final reader combines the initial encoding and completed groups before
prediction. This retains grouped update contributions rather than every
full hidden state; \cref{app:implementation} gives the recurrence.

\begin{figure}[!t]
  \centering
  \includegraphics[width=\linewidth]{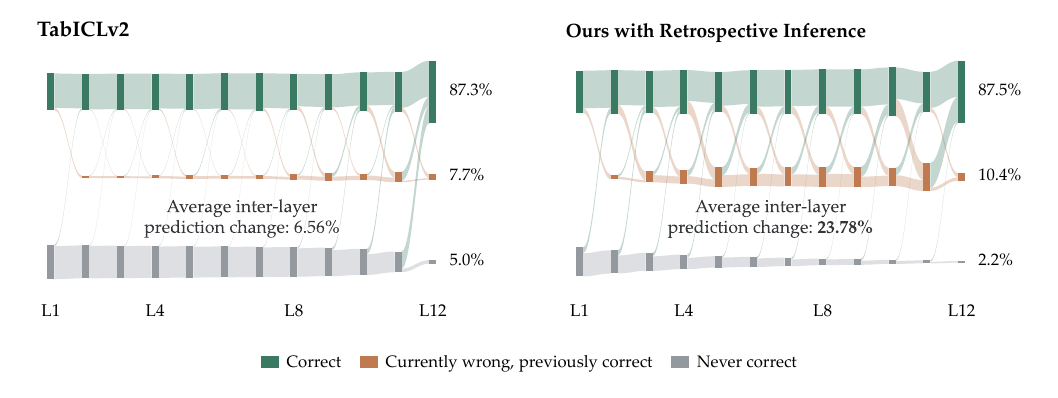}
  \vspace{-20pt}
  \caption{\textbf{Sample-state flows across depth.} Frozen-readout trajectories for TabICLv2 (left) and our ungated retrospective variant (right), averaged equally over 38 classification datasets. \emph{Correct} merges always-correct and previously-wrong but currently-correct queries; the two wrong states distinguish whether a query was ever correct. Ribbon widths show query fractions across all 12 layers. The displayed average measures correctness switches over L1$\to$L2 through L10$\to$L11; only this summary excludes the final transition.}
  \label{fig:retrospective-dynamics}
\end{figure}

\noindent{\bf Representation dynamics.}
We examine the resulting trajectories with the fixed final-support readout
used in \cref{sec:dynamics}. In \cref{fig:representation-comparison}
(right), predictive changes appear earlier and involve different queries
at successive stages. The aggregate state flows in
\cref{fig:retrospective-dynamics} show revisions distributed across depth,
including both corrections of previously incorrect predictions and
deteriorations of previously correct ones.\looseness=-1

Under this fixed readout, retrospective inference exhibits earlier and more distributed prediction changes, with successive stages revising different subsets of queries. This pattern is consistent with a form of multi-view refinement, where different stages contribute to different subsets of queries.
Additional model comparisons and classification and regression cases are reported in \cref{fig:sdss17-state-flows,fig:benchmark-state-flows,app:regression-views,app:classification-views}.

\subsection{Shaping Contextual Updates with Gated Attention}
\label{sec:selective}

Historical reads determine the input to a sublayer; its output becomes a
new contribution available for later reuse. We use Gated Attention
\citep{qiu2025gated} to modulate this contextual update for each receiving
row. Let $z_{\ell i}$ be the normalized attention-reader output and
$o_{\ell i}$ the concatenated attention-head outputs. Following the
row-vector convention of \cref{eq:icl-attention},
\begin{equation}
\begin{aligned}
g_{\ell i}
  &=\sigma(z_{\ell i}W_{g,\ell}^{\top}+\mathbf b_{g,\ell}^{\top}),\\
A_{\ell i}
  &=(g_{\ell i}\odot o_{\ell i})W_{O,\ell}^{\top}
    +\mathbf b_{O,\ell}^{\top}.
\end{aligned}
\label{eq:gate}
\end{equation}
Here $\sigma$ is the sigmoid. Each layer learns a gate projection shared
across support and query rows. Ordinary attention is already
query-dependent; gating additionally modulates its output channels before
projection. It can therefore alter both the direction and magnitude of
the update, rather than only its overall strength. The gated update enters
the current group's accumulation, shaping the information retained for
subsequent reads.

\noindent{\bf Direction and magnitude.}
We analyze fixed gated models on 38 classification and 13 regression
datasets, using the first official split, one view without ensembling,
and native final prediction heads. A counterpart update replaces each
query's gate with the channelwise mean over the unlabeled query batch.
We substitute either this full update, its direction while retaining the
learned norm, or its norm while retaining the learned direction.
Replacements occur after output projection at every ICL layer, on query
rows only. All normalization layers remain in place; the updates are
recomputed at each intervened state (\cref{app:gating-geometry}).

\begin{figure}[!t]
  \centering
  \begin{minipage}[t]{0.485\linewidth}
    \vspace{0pt}
    \centering
    \includegraphics[width=\linewidth]{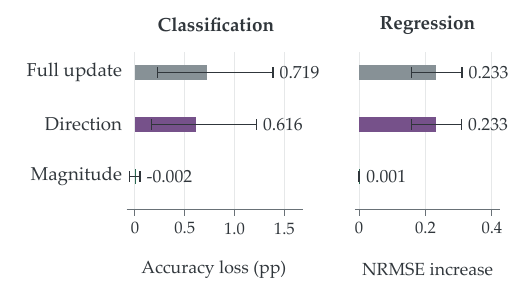}
    \caption{\textbf{Gated-update geometry.} Mean-gate replacement effects on 38 classification and 13 regression datasets in TabArena. Whiskers: 95\% dataset-bootstrap intervals.}
    \label{fig:gated-incorporation}
  \end{minipage}\hfill
  \begin{minipage}[t]{0.485\linewidth}
    \vspace{0pt}
    \centering
    \includegraphics[width=\linewidth]{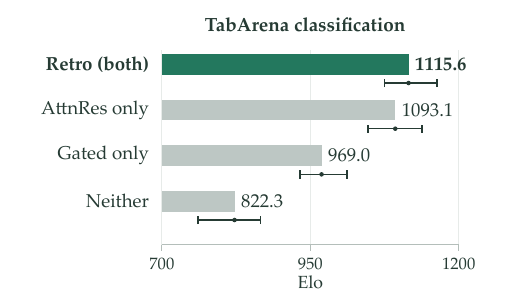}
    \caption{\textbf{Component ablation.} TabArena classification Elo within the four-configuration comparison. Whiskers: reported 95\% intervals.}
    \label{fig:component-ablation}
  \end{minipage}
\end{figure}

Replacing direction reduces mean classification accuracy by 0.616 percentage points, compared with $-0.002$ for replacing magnitude (\cref{fig:gated-incorporation}); the paired difference is 0.618 points (95\% interval: [0.173, 1.217]). The corresponding regression NRMSE increases are 0.2328 and 0.0011, with NRMSE defined as query RMSE divided by the support-target standard deviation. 
Thus, under these replacements, the trained models depend more on the learned update direction than on its norm. This supports a role for gating in shaping contextual information, beyond attenuating its strength. \Cref{fig:component-ablation} provides the complementary component ablation, with dataset-level results and additional controls reported in \cref{app:gating-discussion}.

\section{\Retro{}: Model, Training, and Evaluation}
\label{sec:retro}

We combine retrospective access and selective contextual incorporation in \Retro{}, a tabular foundation model built on a
hierarchical table encoder and a row-level ICL stack. We describe its architecture and pretraining procedure, then evaluate its predictive performance, component contributions, and computational cost.

\begin{figure}[!t]
  \centering
  \includegraphics[width=\linewidth]{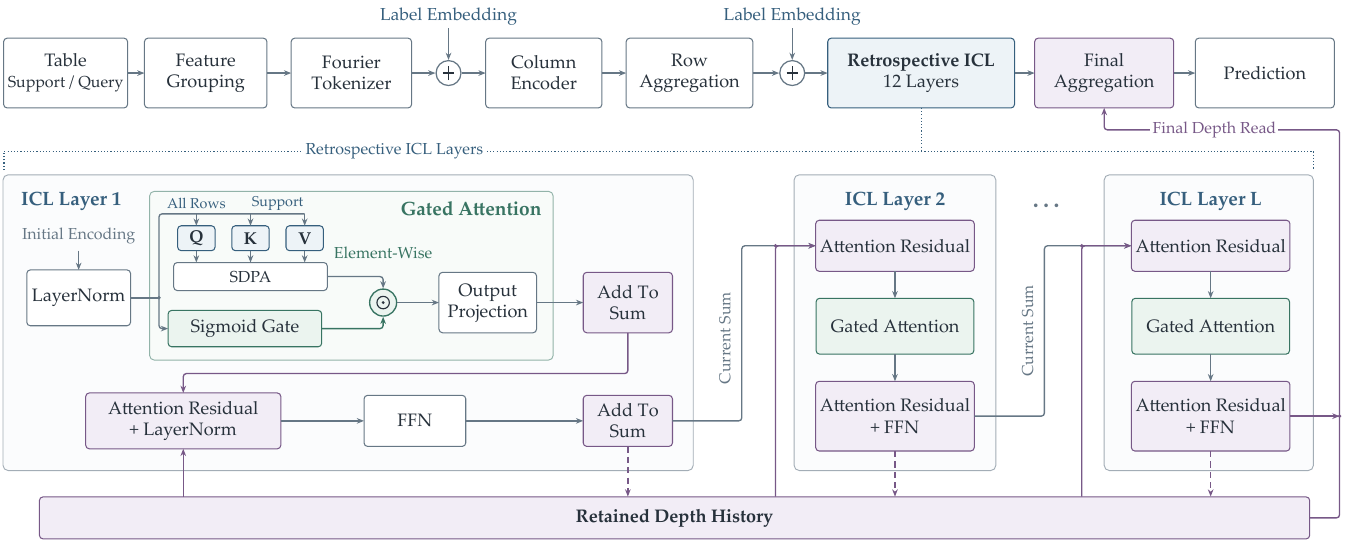}
  \caption{\textbf{The \Retro{} architecture.} A table encoder produces row representations for a retrospective ICL stack.
  Within each layer, separate aggregation modules construct attention and feed-forward inputs from available historical contributions. An input-conditioned gate modulates attention outputs before projection. Completed two-layer update groups remain available to subsequent layers and the final aggregation used for prediction.}
  \label{fig:retro-architecture}
\end{figure}

\noindent{\bf Architecture.}
As illustrated in \cref{fig:retro-architecture}, the table encoder consists of three column-processing blocks and three row-processing blocks, both with a hidden dimension of 128 and eight attention heads. The column stage uses 128 inducing vectors, while the row stage aggregates feature representations through four CLS tokens. Their outputs are concatenated into a 512-dimensional representation for each row. We retain TabICLv2's repeated feature grouping, target-aware embeddings, and query-aware scalable softmax (QASSMax).
The row representations, augmented with observed support labels, enter a 12-layer ICL stack with
eight attention heads and a feed-forward expansion ratio of two. Within this stack, retrospective aggregation constructs
the inputs to attention and feed-forward sublayers, and gated attention modulates contextual outputs
before projection (\cref{eq:attnres,eq:gate}). Residual updates are grouped over pairs of layers.
The final aggregation combines the initial encoding and six completed update groups, and a two-layer MLP
with hidden dimension 1,024 produces query predictions. Following TabICLv2, classification and regression use
separately trained models with task-specific target embeddings and output heads.

\noindent{\bf Pretraining.}
We use the synthetic task generator of TabICLv2, which samples structural causal models with diverse graph structures and functional relationships to generate heterogeneous numerical and categorical features. Each task is divided into labeled support examples and held-out queries. The classification model learns to predict query class labels, while the regression model predicts 999 quantiles at levels $\{0.001,0.002,\ldots,0.999\}$ using the summed
pinball loss.
Pretraining follows a three-stage curriculum that progressively increases the number of samples per
task, with up to 100 features and a batch size of 64 tasks. The first stage uses 500K updates on tasks with 1,024 samples, allocating 30--90\% to the support set. The second and third stages use 40K and 10K updates, respectively, with task sizes sampled log-uniformly from 400--10,240 and 400--60,000 samples and an 80\% support fraction.
We use Muon for matrix-valued parameters and AdamW for one-dimensional parameters, following the optimization setup of TabICLv2. We adopt cosine learning-rate schedules with peak rates of $8\times10^{-4}$, $10^{-4}$, and $2\times10^{-5}$ across the three stages, and a weight decay of 0.01. Further details of synthetic data generation, filtering, and the remaining optimization settings follow~\citet{qu2026tabiclv2}.

\noindent{\bf Evaluation protocol and predictive performance.}
At inference time, pretrained parameters remain fixed and labeled support examples provide task-specific context. Classification predictions are obtained from the classification head; regression point predictions are computed by averaging the predicted quantiles. We retain the preprocessing and inference procedures of TabICLv2. The benchmark scope, available evaluation settings, and aggregation procedures are described in \cref{app:benchmark-results}.

We evaluate on TabArena, TALENT, and RelArena using the comparison pools listed in \cref{app:benchmark-results}. 
We report separate classification and regression Elo rankings, together with overall ratings recomputed from pooled results rather than averaged across the two task types.

\begin{figure}[!t]
  \centering
  \includegraphics[width=\linewidth]{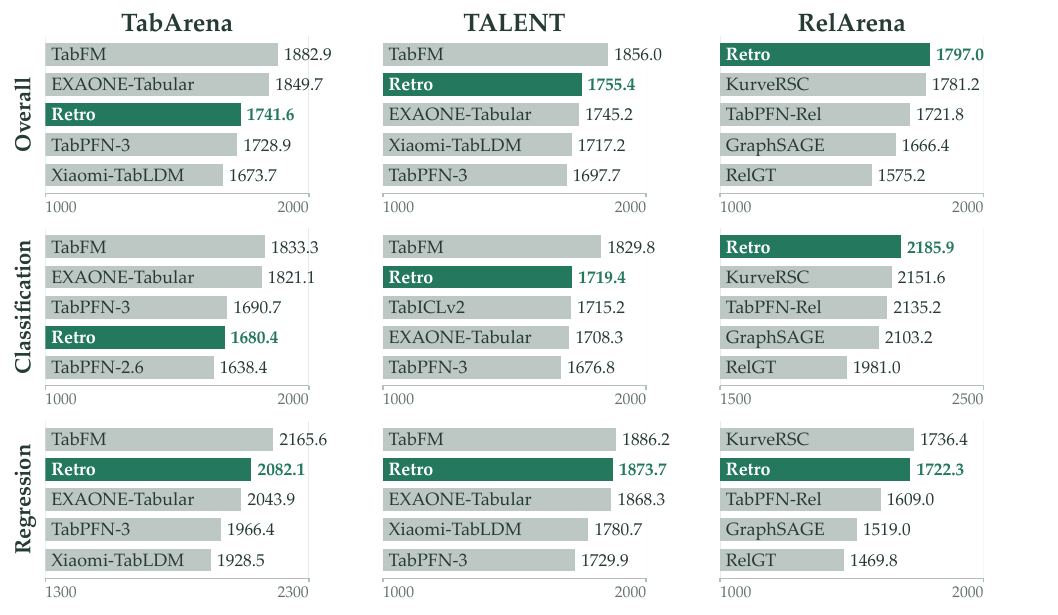}
  \vspace{-15pt}
  \caption{\textbf{Elo rankings across three benchmarks.} Each panel shows the five highest-rated methods. Candidate sets and scoring protocols differ, so ratings should be compared within panels. Complete rankings appear in \cref{app:benchmark-results}.}
  \label{fig:benchmark-elo}
\end{figure}

\Cref{fig:benchmark-elo} shows the five highest-rated methods in each ranking. Among the evaluated methods, \Retro{} ranks third on TabArena, second on TALENT, and first on RelArena by overall Elo. It also ranks above TabICLv2 in the overall rankings on TabArena and TALENT, indicating competitive performance relative to the backbone on which our architecture is based.

The task-specific rankings reveal variation across evaluation settings.
On TabArena, \Retro{} ranks fourth for classification and second for regression.
On TALENT, it ranks second for both task types, although its overall Elo is close to those of several competing TFMs.
On RelArena, it ranks first for classification and second for regression.
These results establish competitive performance across classification and regression, while the relative advantages depend on the benchmark. Complete rankings and uncertainty estimates are provided in \cref{app:benchmark-results}.

\noindent{\bf Component ablation.}
On TabArena classification (\cref{fig:component-ablation}), the combined
model has the highest Elo point estimate (1115.6) in this ablation,
while both single-mechanism variants score above the model with neither.
These Elo values use a separate comparison pool from the main rankings
(\cref{app:component-comparison}).

\noindent{\bf Efficiency.}
\Retro{} retains the backbone's width and depth, adding only historical aggregation modules and gate projections within the ICL stack. The aggregation modules introduce 24,576 parameters. With a $d$-dimensional gate in each of the $L$ layers,
gate projections add $L(d^2+d)$ parameters, approximately 3.15M for $L=12$ and $d=512$.
For $n$ rows and $B$ available historical sources, each aggregation requires $\mathcal{O}(Bnd)$ computation, while each gate projection requires $\mathcal{O}(nd^2)$. These operations complement the existing $\mathcal{O}(nn_s d)$ support-conditioned attention and $\mathcal{O}(nd^2)$ feed-forward computation, where $n_s$ is the support-set size.
The performance--runtime trade-off in~\cref{fig:benchmark-pareto} shows that \Retro{} improves predictive performance over TabICLv2 with nearly unchanged inference time.
Relative to the other evaluated methods, it lies on the Pareto frontier: no compared method achieves both better predictive performance and lower runtime under the reported evaluation setting.
This positions \Retro{} as a competitive choice when both predictive quality and inference cost matter. \looseness=-1

\section{Discussion and Conclusion}
\label{sec:conclusion}

We studied how individual queries are refined across the depth of contemporary
TFMs and found that predictive changes are uneven, often
concentrating in later layers. This observation motivates \textit{retrospective
inference}: retaining access to earlier contributions while adapting how
newly computed contextual information modifies each query. \Retro{}
instantiates these two operations through Attention Residuals and
query-conditioned, element-wise Gated Attention.

Retrospective variants exhibit earlier and more distributed prediction
changes, with different stages revising different subsets of queries.
Fixed-weight gating interventions further reveal sensitivity to update
direction beyond changes in magnitude alone. Native-prediction comparisons
place \Retro{} among the top three methods by overall Elo on TabArena,
TALENT, and RelArena within the evaluated candidate sets. Together, these findings support organizing intermediate
computation as reusable information rather than only as transient states
passed from layer to layer.

\clearpage

\bibliographystyle{iclr2027_conference}
\bibliography{ref}
\clearpage

\appendix
\crefalias{section}{appendix}
\crefalias{subsection}{subappendix}
\crefalias{subsubsection}{subsubappendix}

\section{Layer-wise Representation Analysis}
\label{app:protocol}

This appendix specifies the diagnostic used in
\cref{sec:dynamics,sec:retrospection} and reports aggregate depth statistics.
\Cref{app:implementation} describes the retrospective computation, and
\cref{app:figures} presents the full-depth cases. The fixed-readout analysis
here is distinct from the native-head gating interventions in
\cref{app:gating-discussion} and the benchmark evaluations in
\cref{app:benchmark-results}.

\subsection{Data and query-role representations}
\label{app:extraction}
We analyze the first official repeat/fold ($r0/f0$) of each of the
38 TabArena classification datasets \citep{erickson2025tabarena}.
Models use the same outer query rows, a single preprocessing view, and
no ensemble.

Following \citet{ye2025closer}, support representations are obtained by
cross-fitting within the outer support set. Each held-out support fold is
encoded as unlabeled queries conditioned on the remaining support rows.
The number of stratified folds is $\min(10,n_{\min})$, where $n_{\min}$
is the smallest support-class count; the random seed is 42.
Outer queries are encoded against the full outer support set.
Their labels enter neither representation extraction nor readout fitting.
Cross-fitting removes the support row's own label from its encoding,
although the cross-fitted and outer-query contexts have different sizes.

\paragraph{Capture points.}
For TabICLv2, each intermediate representation is the input to the next
block's pre-attention normalization, with the final state taken from the
encoder output. For the retrospective variants, this point includes
the next attention reader; the final state includes the output reader.
It is therefore the accessible mixture, not the raw update accumulator.
Both sequences contain 12 states of width 512.
TabPFN-3 supplies 24 ICL block outputs of width 512 before final RMS
normalization. EXAONE-Tabular supplies 12 target-cell CAST outputs of
width 192 after block normalization. A depth index denotes a native block,
not equal computation across models.
The retrospective diagnostics use Attention Residuals without gating.

\subsection{Fixed readouts and predictive changes}
\label{app:readout}
A StandardScaler and L2-regularized logistic classifier are fitted to the
final-layer cross-fitted support representations. We use $C=1$ and LBFGS
with tolerance $10^{-6}$, allowing up to 20,000 iterations when the initial
3,000-iteration limit is insufficient for convergence. The scaler and
classifier remain fixed across depth.

Let $r_L$ denote this readout and $p_{\ell i}=r_L(h_{\ell i})$.
For query $i$ with true class $y_i^q$, define
\begin{equation}
 m_{\ell i}=p_{\ell i,y_i^q}-\max_{c\ne y_i^q}p_{\ell i,c},
 \qquad
 c_{\ell i}=\mathbf 1\!\left[\arg\max_c p_{\ell i,c}=y_i^q\right].
 \label{eq:margin}
\end{equation}
The probability margin and correctness indicator provide a common decision
reference across depth. They are not predictions from the pretrained
model's native head.

\paragraph{Transitions.}
At layer $\ell>1$, correction and deterioration fractions are
$C_\ell=\E_i[(1-c_{\ell-1,i})c_{\ell i}]$ and
$D_\ell=\E_i[c_{\ell-1,i}(1-c_{\ell i})]$.
Writing $a_\ell=\E_i[c_{\ell i}]$ gives
\begin{equation}
 a_\ell-a_{\ell-1}=C_\ell-D_\ell,\qquad
 R_\ell=C_\ell+D_\ell,\qquad
 V_\ell=\E_i|m_{\ell i}-m_{\ell-1,i}|.
 \label{eq:revision}
\end{equation}
A correctness switch does not include a change between two wrong classes.
The early-motion share is
$\sum_{\ell=2}^{\lfloor2L/3\rfloor}V_\ell/
 \sum_{\ell=2}^{L}V_\ell$; its complement is the final-third share in
\cref{sec:dynamics}. The early range contains seven of eleven transitions
for 12-layer models and fifteen of twenty-three for TabPFN-3.
All query rows contribute to each dataset's statistics, followed by an
equal-weight average across datasets.

State-flow plots use three histories: \emph{Correct},
\emph{Currently wrong, previously correct}, and \emph{Never correct}.
The second refers to any earlier correct state, not only the preceding
layer; the third includes the current layer.
The two currently-correct histories of the underlying four-state
partition are merged by summing their node counts and transition counts.
Every query and native layer is retained.
The scalar summary printed on the flow plots averages $R_\ell$ over
$\ell=2,\ldots,L-1$, excluding the final transition.
In contrast, the per-step statistics in \cref{tab:depth-dynamics} include
all $L-1$ transitions.

\subsection{Visualization and regression}
\label{app:visualization}
For each model and dataset, PCA is fitted to at most 1,200 deterministically
selected final-layer support embeddings. Up to 1,200 query rows are shown.
Every layer uses the same basis and axis limits within a model; coordinates
are not aligned across models. Readouts operate in the original embedding
space, so the plots are not two-dimensional decision surfaces.

In classification plots, green and purple denote positive and negative
adjacent margin changes. Rings mark corrections and crosses mark
deteriorations. The first measured layer is gray because it has no measured
predecessor. The main SDSS17 figure includes separately labeled inset
scales to resolve small early-layer changes. The remaining classification
plates share the same main color scale.

\paragraph{Regression.}
The superconductivity case in \cref{app:regression-views} contains 14,175
support and 7,088 query rows. Support extraction uses shuffled ten-fold
cross-fitting with seed 42. A StandardScaler and Ridge regressor
($\alpha=1$, with intercept) are fitted to final-layer cross-fitted
support representations and frozen across depth. Colors encode
\begin{equation}
 \delta e_{\ell i}
 =\frac{|\widehat y_{\ell-1,i}-y_i^q|
          -|\widehat y_{\ell i}-y_i^q|}{s_S},
 \label{eq:regression-change}
\end{equation}
where $s_S$ is the outer-support target standard deviation.
Positive values indicate error reduction. The four models share a
symmetric-log scale spanning $[-10,10]$; no classification-state partition
is assigned to regression.

\subsection{Quantitative depth diagnostics}
\label{app:depth-dynamics}
\Cref{tab:depth-dynamics} summarizes the ungated retrospective variant
used in \cref{fig:representation-comparison}.
Mean correctness turnover per transition is 24.66\% for the retrospective
variant and 9.82\% for TabICLv2; 79.5\% and 63.8\% of queries, respectively,
change correctness at least once. The early-motion shares are 60.7\%
and 15.6\%, with the retrospective share larger on 36 of 38 datasets.
Turnover and absolute margin movement are larger on 33 of 38 datasets.
In the state-flow comparison in \cref{fig:retrospective-dynamics}, final
frozen-readout accuracies are 87.3\% for TabICLv2 and 87.5\% for our
retrospective model.

\begin{table}[t]
  \centering
  \setlength{\tabcolsep}{7.8pt}
  \caption{\textbf{Descriptive frozen-readout dynamics, equally averaged over 38 datasets (one official split each).} Bold marks the largest value in each column. Early motion is the share of absolute probability-margin change before the final third of depth. TabPFN-3 has 24 layers; the other models have 12. Native-step averages complement the depth-dependent any-revision statistic, but do not equate compute or depth granularity across architectures.}
  \label{tab:depth-dynamics}
  \vspace{5pt}
  \begin{tabular}{lcccc}
  \toprule
  Model & Any revision (\%) & Revision / step (\%) & $\mathbb{E}|\Delta m|$ & Early motion (\%) \\
  \midrule
  TabICLv2 & 63.8 & 9.82 & 0.197 & 15.6 \\
  TabPFN-3 & 66.8 & 7.55 & 0.149 & 48.5 \\
  EXAONE-Tabular & 71.8 & 16.12 & 0.319 & 38.9 \\
  \midrule
  \textbf{Ours} & \textbf{79.5} & \textbf{24.66} & \textbf{0.488} & \textbf{60.7} \\
  \bottomrule
  \end{tabular}
\end{table}

The per-step statistics here include all transitions, whereas the scalar
summaries in \cref{fig:benchmark-state-flows} exclude the final transition.
The model settings for each analysis are specified in \cref{app:extraction}.
The case studies in \cref{app:figures} complement both summaries by locating
changes within individual query trajectories.

\subsection{Interpretation}
\label{app:diagnostic-limits}
The frozen-readout analysis measures how predictive information is expressed relative to a common final-layer decision rule, rather than the total information contained in each representation. We therefore use it to compare layer-wise refinement patterns; the aggregate statistics include all 38 classification datasets, while the visual cases are illustrative.
\section{Retrospective ICL Implementation}
\label{app:implementation}

This appendix specifies the grouped accumulation underlying
\cref{eq:attnres,eq:gate} and the architecture in
\cref{fig:retro-architecture}. The table encoder and pretraining procedure
are described in \cref{sec:retro}.

\subsection{Historical reads and grouped updates}
\label{app:grouped-history}
Let $P$ be the current group accumulator and $M$ the ordered bank of
completed sources. Initially $P=H_0$ and $M$ is empty.
For layer $\ell$, the attention input is
\begin{equation}
 U_\ell=
 \begin{cases}
  P, & \ell=1,\\
  \operatorname{Read}_{a,\ell}(M,P), & \ell>1,
 \end{cases}
 \label{eq:grouped-attention-read}
\end{equation}
where each $\operatorname{Read}$ applies \cref{eq:attnres} independently
to every row. At the start of a two-layer group
($\ell=1,3,\ldots,11$), this read is followed by appending $P$ to $M$
and resetting $P$ to zero. Thus, the initial encoding becomes the first
stored source. Attention and feed-forward updates then proceed as
\begin{align}
 P &\leftarrow P+A_\ell(\operatorname{LN}(U_\ell);S),\nonumber\\
 V_\ell&=\operatorname{Read}_{f,\ell}(M,P),\nonumber\\
 P &\leftarrow P+\operatorname{FFN}_\ell(\operatorname{LN}(V_\ell)).
 \label{eq:grouped-updates}
\end{align}
Here $A_\ell$ includes the attention output projection and, when enabled,
the gate in \cref{eq:gate}. Both support and query rows receive updates,
but only support rows supply attention keys and values.
The attention read occurs before a group is stored; the feed-forward
read occurs after the attention update is added.

After the last layer, an independent output reader combines $(M,P)$.
For 12 layers grouped in pairs, it reads seven sources: the initial
encoding and six group sums. The capture points in \cref{app:extraction}
include these learned reads rather than only the accumulator.

\subsection{Normalization, initialization, and gating}
Each reader has its own length-$d$ scoring vector and RMS gain, shared
across rows. The scorer is initialized to zero and the gain to one, giving
uniform source weights initially. RMS normalization, scoring, softmax,
and weighted reduction are evaluated in FP32; the result is cast to
the input dtype. Only scoring uses normalized sources.
The attention and feed-forward sublayers retain their pre-LayerNorms.

The scorer is a learned site-specific vector, not a projection of the
latest state. Consequently, weights vary with the row's source values
and the reader's parameters, without inheriting a previous reader's weights.
Gating is computed from the normalized attention input and multiplies
the concatenated head outputs before their output projection.
It modulates latent channels but does not skip attention or feed-forward
computation. Its experimental role is examined in
\cref{app:gating-discussion}.

There are $(L-1)+L+1=2L$ historical readers, adding $4Ld$ parameters,
or 24,576 at $L=12,d=512$. The $L$ gate projections add $L(d^2+d)$
parameters. These counts cover the additional modules, not the total
model size or measured runtime.

\clearpage
\section{Full-Depth Representation and State-Flow Analyses}
\label{app:figures}
The following case studies extend \cref{sec:dynamics,sec:retrospection}
using the extraction and readout protocol in \cref{app:extraction,app:readout}.
We first compare full-depth SDSS17 representations and state flows, then
report benchmark-average flows and additional regression and classification
cases. All measured layers are retained. Within each model, every layer
uses the final-support PCA basis; neither coordinates nor computation per
layer are aligned across models. All trajectories use the fixed readouts
defined in \cref{app:readout}, while \cref{app:depth-dynamics} reports
aggregate classification statistics.

\makeatletter
\newenvironment{appendixfigure}{\par\noindent\begin{minipage}{\linewidth}\def\@captype{figure}\centering}{\end{minipage}\par}
\makeatother

\subsection{SDSS17: full-depth representations}
\label{app:baseline-views}
\Crefrange{fig:sdss17-tabiclv2}{fig:sdss17-exaone} extend \cref{fig:representation-comparison} to all four models on the same SDSS17 split. TabPFN-3 has 24 layers; the other models have 12. These full-depth views reveal when predictive changes emerge and which queries they affect, complementing the comparison of early and late refinement in \cref{sec:dynamics}.
\begin{appendixfigure}
\includegraphics[width=\linewidth,height=0.78\textheight,keepaspectratio]{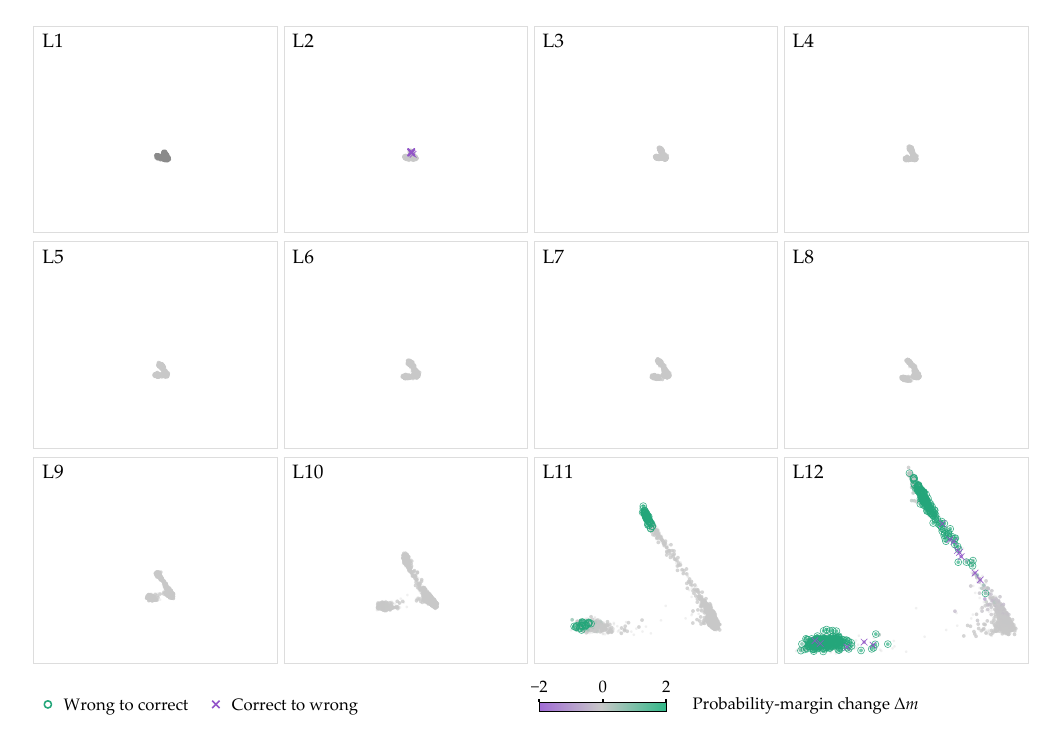}
\caption{\textbf{TabICLv2 on SDSS17.} All native layers in the fixed final-support PCA basis. Colors show adjacent probability-margin change under the frozen readout; rings and crosses mark corrections and deteriorations. The first layer is gray. Extraction and display conventions are given in \cref{app:extraction,app:visualization}.}\label{fig:sdss17-tabiclv2}
\end{appendixfigure}\clearpage
\begin{appendixfigure}
\includegraphics[width=\linewidth,height=0.78\textheight,keepaspectratio]{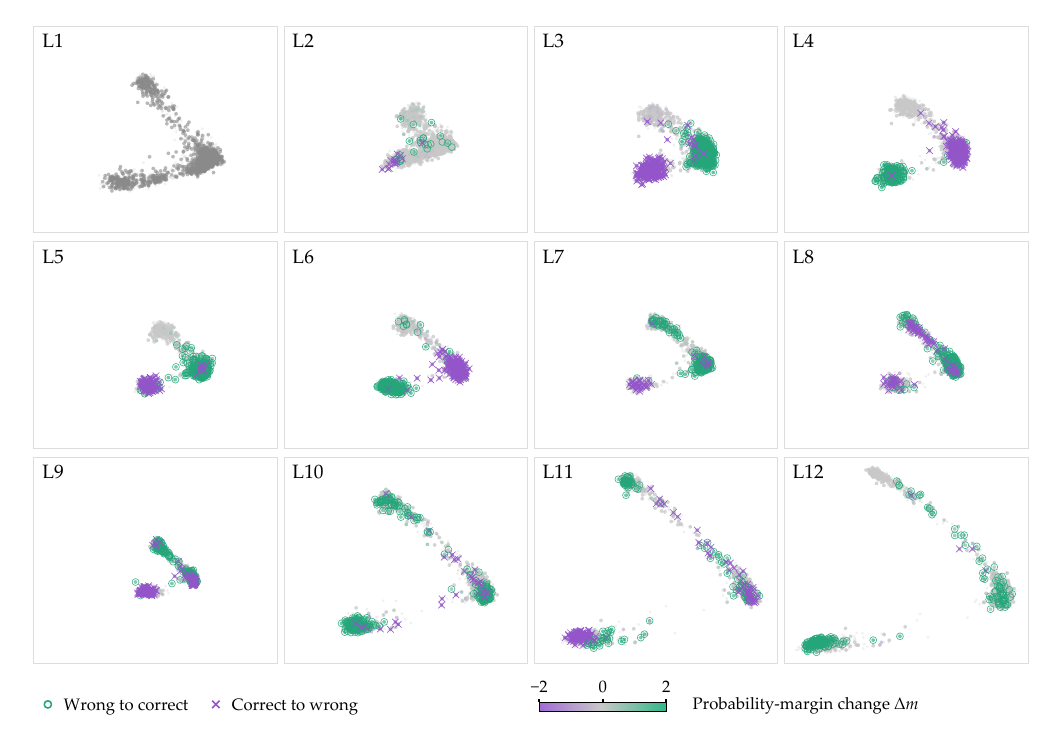}
\caption{\textbf{Our retrospective model on SDSS17.} All native layers in the fixed final-support PCA basis. Colors show adjacent probability-margin change under the frozen readout; rings and crosses mark corrections and deteriorations. The first layer is gray. Extraction and display conventions are given in \cref{app:extraction,app:visualization}.}\label{fig:sdss17-attnres}
\end{appendixfigure}\clearpage
\begin{appendixfigure}
\includegraphics[width=\linewidth,height=0.78\textheight,keepaspectratio]{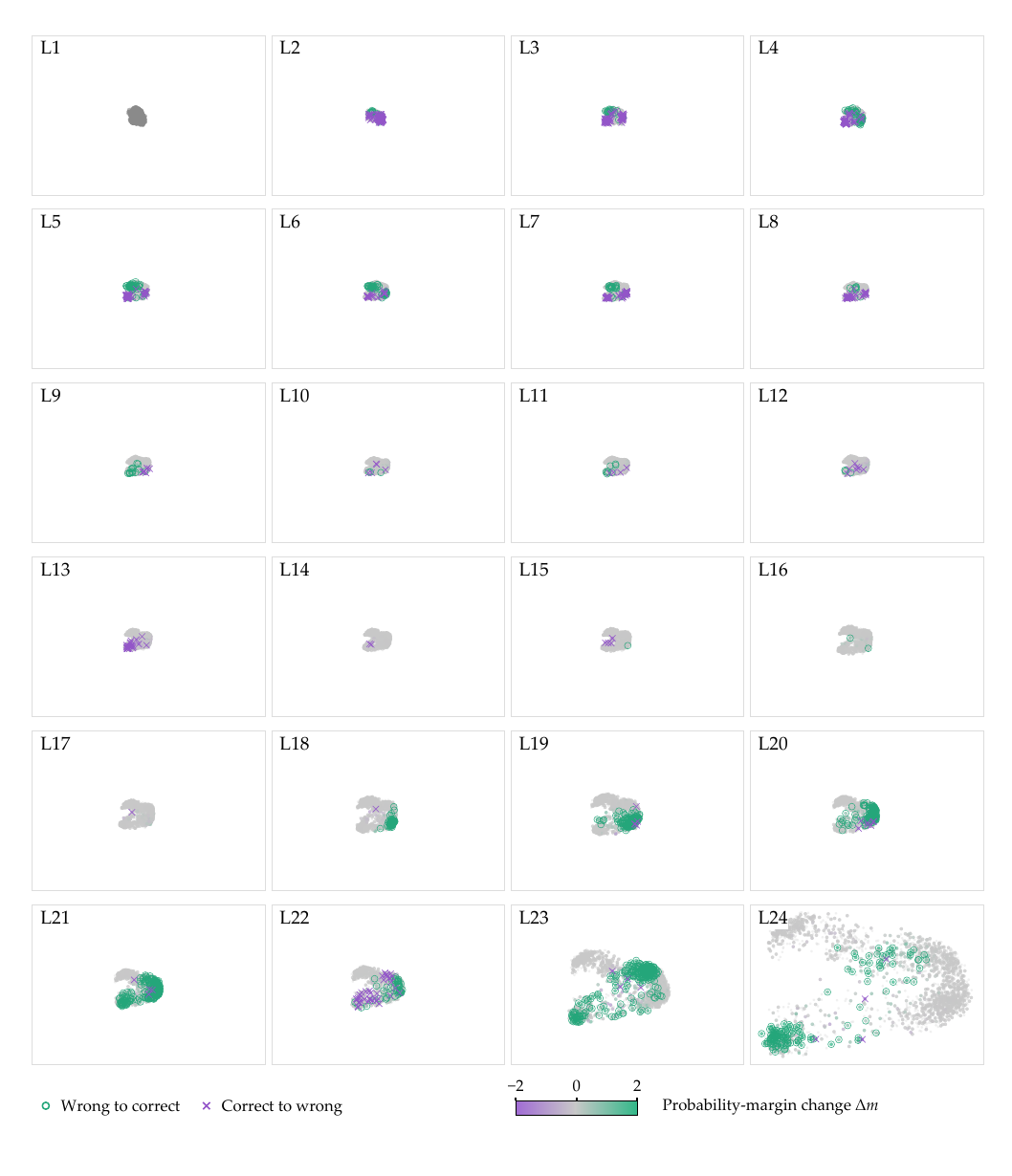}
\caption{\textbf{TabPFN-3 on SDSS17.} All native layers in the fixed final-support PCA basis. Colors show adjacent probability-margin change under the frozen readout; rings and crosses mark corrections and deteriorations. The first layer is gray. Extraction and display conventions are given in \cref{app:extraction,app:visualization}.}\label{fig:sdss17-tabpfn3}\label{fig:appendix-baselines}
\end{appendixfigure}\clearpage
\begin{appendixfigure}
\includegraphics[width=\linewidth,height=0.78\textheight,keepaspectratio]{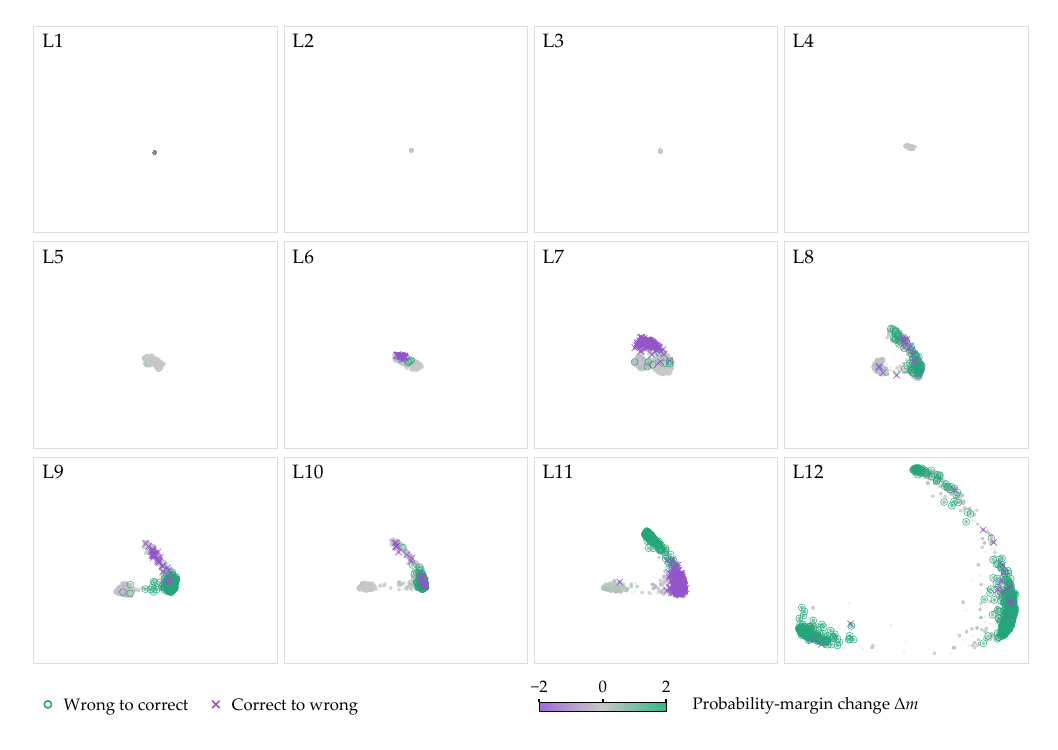}
\caption{\textbf{EXAONE-Tabular on SDSS17.} All native layers in the fixed final-support PCA basis. Colors show adjacent probability-margin change under the frozen readout; rings and crosses mark corrections and deteriorations. The first layer is gray. Extraction and display conventions are given in \cref{app:extraction,app:visualization}.}\label{fig:sdss17-exaone}
\end{appendixfigure}\clearpage

\subsection{SDSS17: sample-state flows}
\label{app:sdss17-flows}
\Cref{fig:sdss17-state-flows} tracks all 26,018 SDSS17 queries using the three-state partition in \cref{app:readout}. Unlike the sampled PCA displays, these flows retain every query. They show how corrections of previously incorrect predictions and deteriorations of previously correct ones contribute to the changing query populations across depth.
\begin{appendixfigure}
\includegraphics[width=\linewidth,height=0.74\textheight,keepaspectratio]{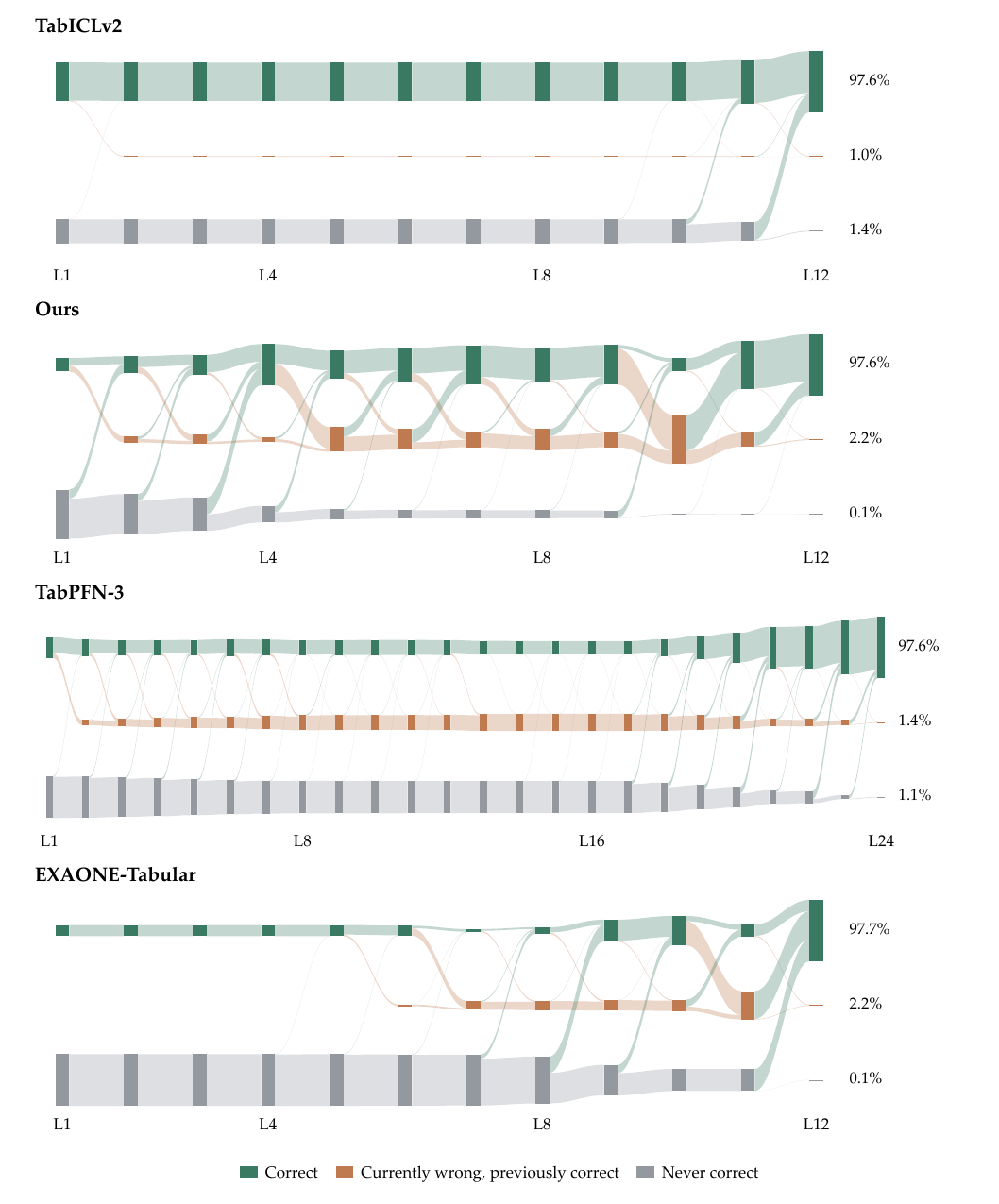}
\caption{\textbf{Four-model sample-state flows on SDSS17.} Node heights and ribbon widths are percentages of all outer queries. \emph{Correct} merges both currently-correct histories; the two wrong states distinguish whether a query was ever correct at an earlier layer. All native layers are retained.}\label{fig:sdss17-state-flows}
\end{appendixfigure}\clearpage

\subsection{Benchmark-average sample-state flows}
\label{app:benchmark-flows}
\Cref{fig:benchmark-state-flows} extends \cref{fig:retrospective-dynamics} to all four models over the 38 classification datasets. Each dataset contributes equally, regardless of query-set size. These flows compare how correctness changes are distributed across depth beyond a single illustrative dataset. Aggregation conventions are given in \cref{app:readout,app:depth-dynamics}.
\begin{appendixfigure}
\includegraphics[width=\linewidth,height=0.74\textheight,keepaspectratio]{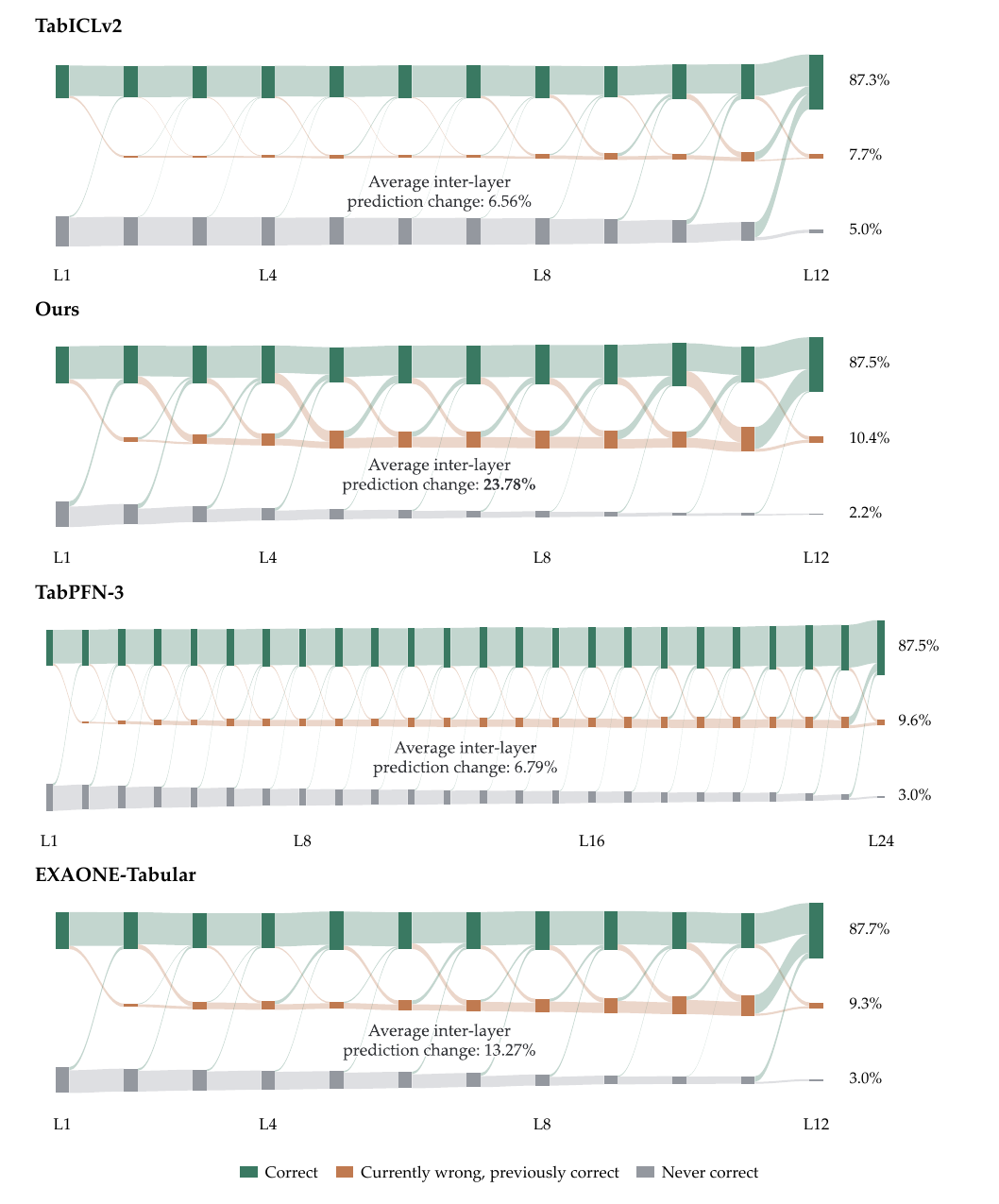}
\caption{\textbf{Four-model sample-state flows averaged over 38 classification datasets.} Node heights and ribbon widths are equal-weight means of per-dataset query percentages, using the same three states as \cref{fig:retrospective-dynamics}. All native layers are shown. The average inter-layer prediction change excludes only each model's final transition from the summary, not from the plotted flows.}\label{fig:benchmark-state-flows}
\end{appendixfigure}\clearpage

\subsection{Superconductivity regression: full-depth representations}
\label{app:regression-views}
\Crefrange{fig:regression-superconductivity-tabiclv2}{fig:regression-superconductivity-exaone} extend the analysis to superconductivity. Colors encode the normalized reduction in absolute error defined in \cref{eq:regression-change}, using one scale across models. This reveals improvements and deteriorations without discretizing a continuous prediction into correctness states.
\begin{appendixfigure}
\includegraphics[width=\linewidth,height=0.78\textheight,keepaspectratio]{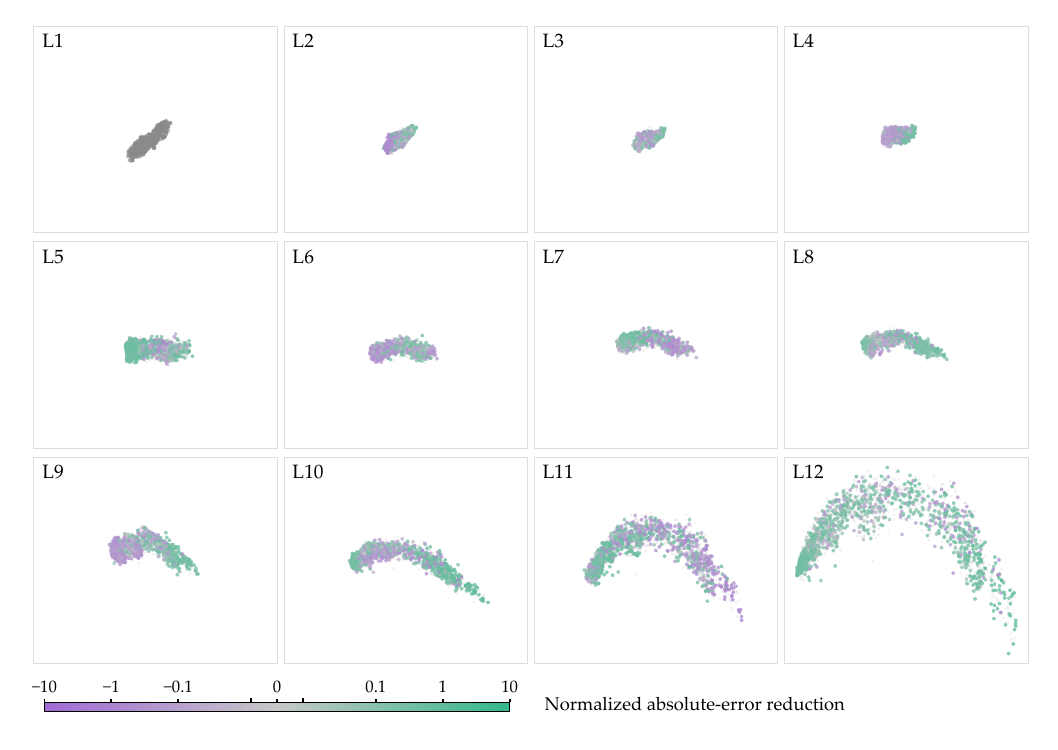}
\caption{\textbf{TabICLv2 on superconductivity.} All native layers in the fixed final-support PCA basis. Green/purple indicate reductions/increases in absolute error under the frozen final-support Ridge readout, normalized by the support-target standard deviation. All four models share the same color scale. The first layer is gray; no binary correctness states are assigned.}\label{fig:regression-superconductivity-tabiclv2}
\end{appendixfigure}\clearpage
\begin{appendixfigure}
\includegraphics[width=\linewidth,height=0.78\textheight,keepaspectratio]{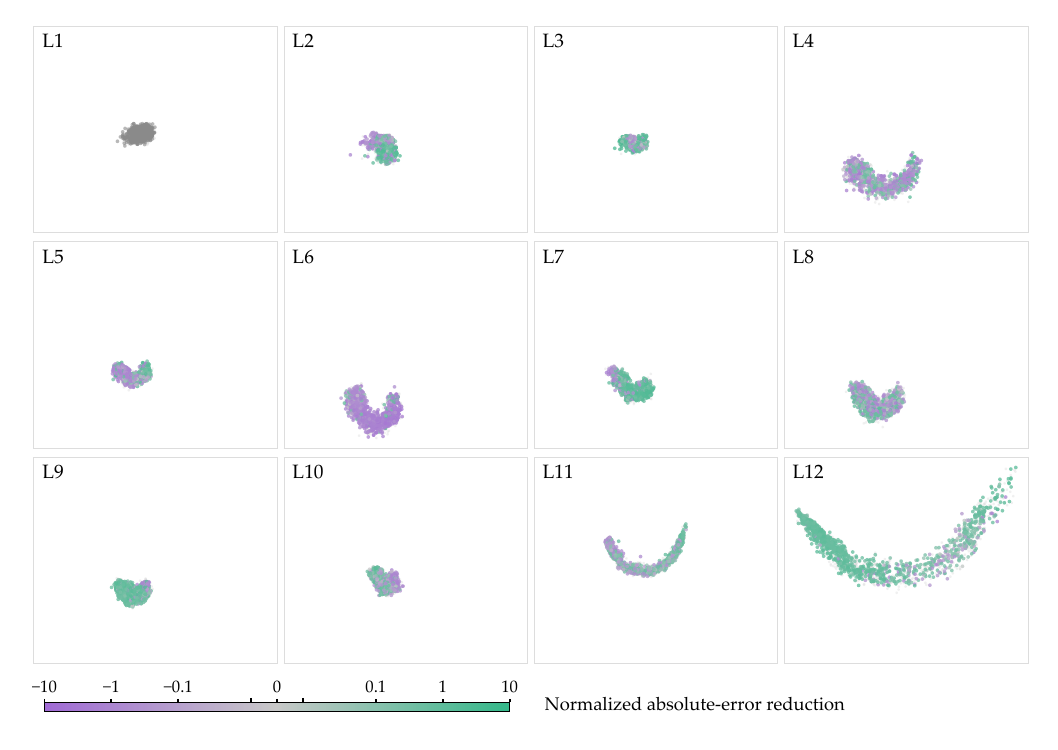}
\caption{\textbf{Our retrospective model on superconductivity.} All native layers in the fixed final-support PCA basis. Green/purple indicate reductions/increases in absolute error under the frozen final-support Ridge readout, normalized by the support-target standard deviation. All four models share the same color scale. The first layer is gray; no binary correctness states are assigned.}\label{fig:regression-superconductivity-attnres}
\end{appendixfigure}\clearpage
\begin{appendixfigure}
\includegraphics[width=\linewidth,height=0.78\textheight,keepaspectratio]{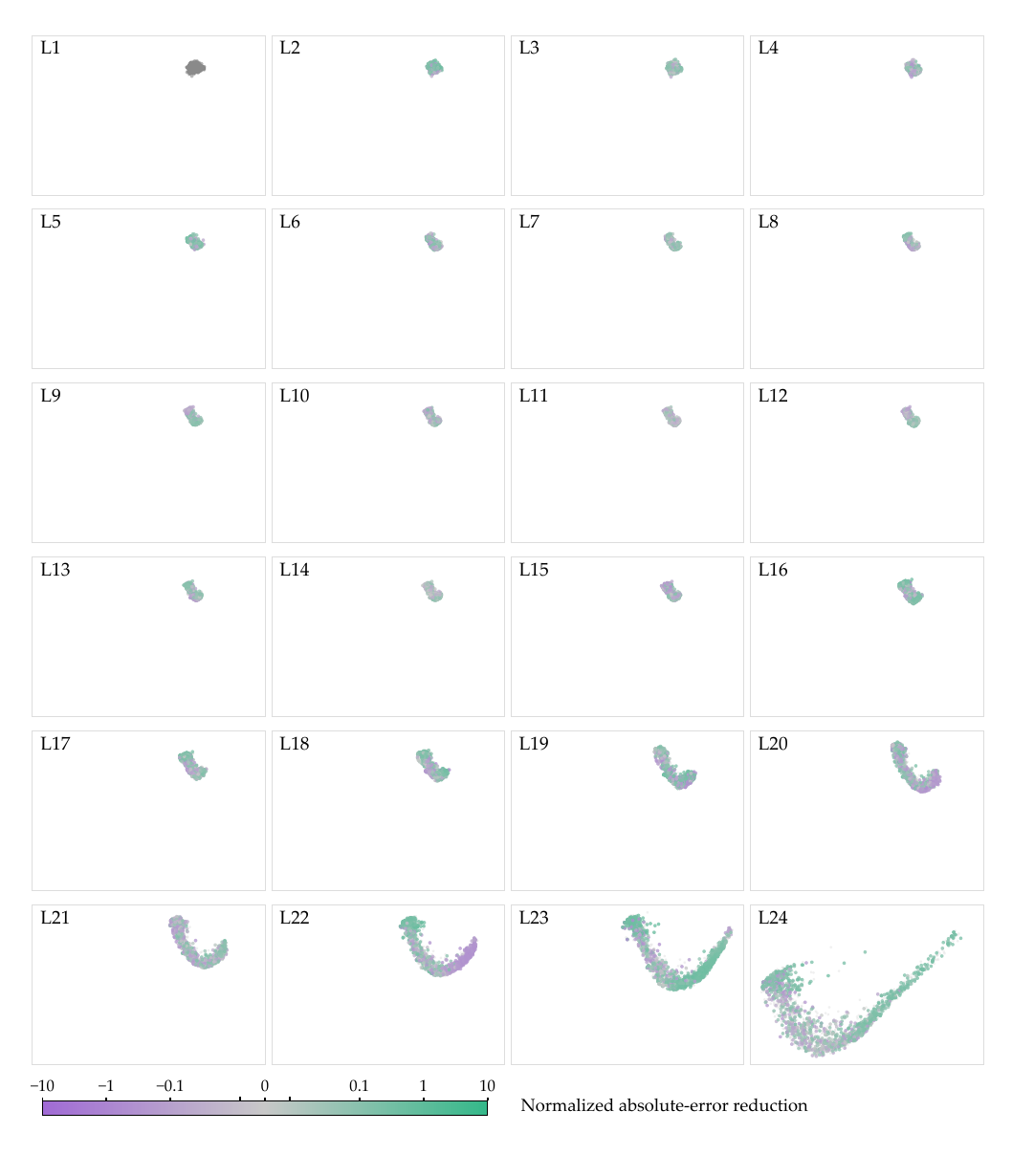}
\caption{\textbf{TabPFN-3 on superconductivity.} All native layers in the fixed final-support PCA basis. Green/purple indicate reductions/increases in absolute error under the frozen final-support Ridge readout, normalized by the support-target standard deviation. All four models share the same color scale. The first layer is gray; no binary correctness states are assigned.}\label{fig:regression-superconductivity-tabpfn3}
\end{appendixfigure}\clearpage
\begin{appendixfigure}
\includegraphics[width=\linewidth,height=0.78\textheight,keepaspectratio]{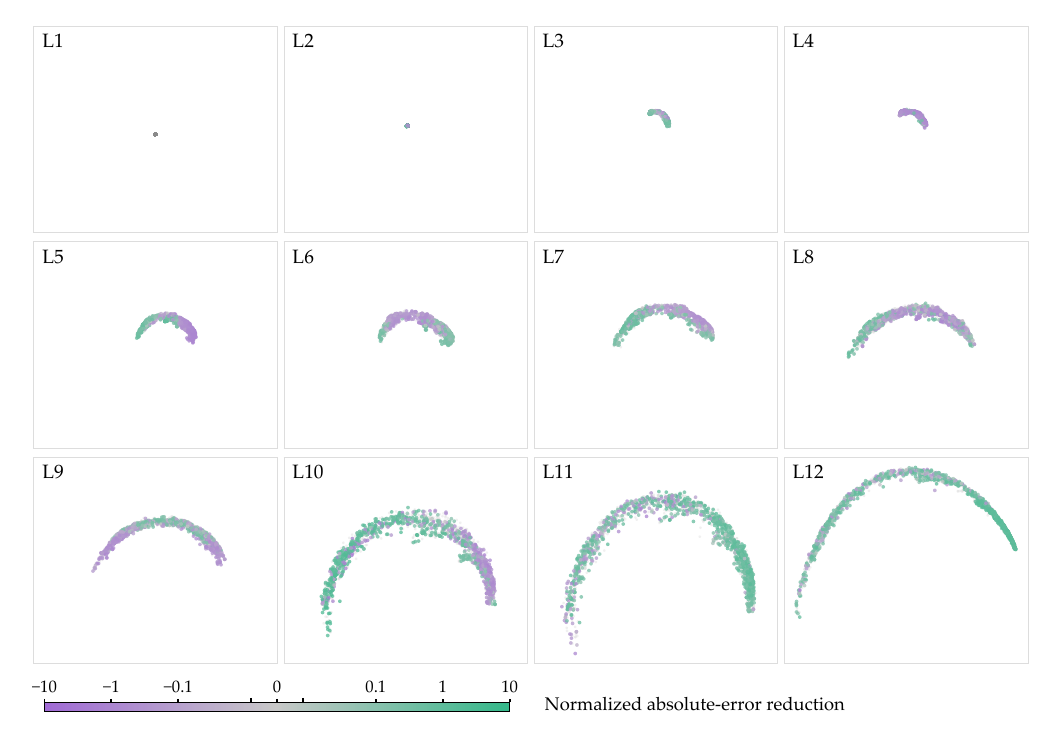}
\caption{\textbf{EXAONE-Tabular on superconductivity.} All native layers in the fixed final-support PCA basis. Green/purple indicate reductions/increases in absolute error under the frozen final-support Ridge readout, normalized by the support-target standard deviation. All four models share the same color scale. The first layer is gray; no binary correctness states are assigned.}\label{fig:regression-superconductivity-exaone}
\end{appendixfigure}\clearpage

\subsection{Additional classification case: Is-this-a-good-customer}
\label{app:classification-views}
\Crefrange{fig:customer-tabiclv2}{fig:customer-exaone} provide an additional classification case. Correct predictions can become incorrect at later layers, illustrating that refinement is not monotonic. These query-level trajectories complement the aggregate state-flow summaries by showing both corrections and deteriorations within the same dataset.
\begin{appendixfigure}
\includegraphics[width=\linewidth,height=0.78\textheight,keepaspectratio]{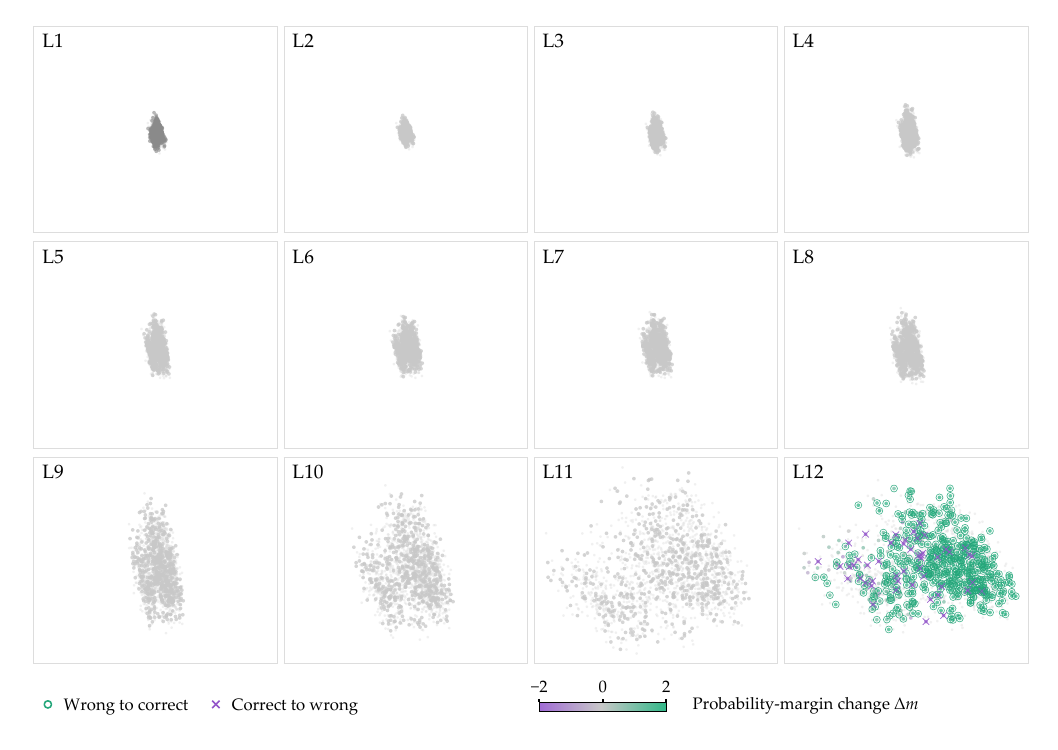}
\caption{\textbf{TabICLv2 on Is-this-a-good-customer.} All native layers in the fixed final-support PCA basis. Colors show adjacent probability-margin change under the frozen readout; rings and crosses mark corrections and deteriorations. The first layer is gray. Extraction and display conventions are given in \cref{app:extraction,app:visualization}.}\label{fig:customer-tabiclv2}
\end{appendixfigure}\clearpage
\begin{appendixfigure}
\includegraphics[width=\linewidth,height=0.78\textheight,keepaspectratio]{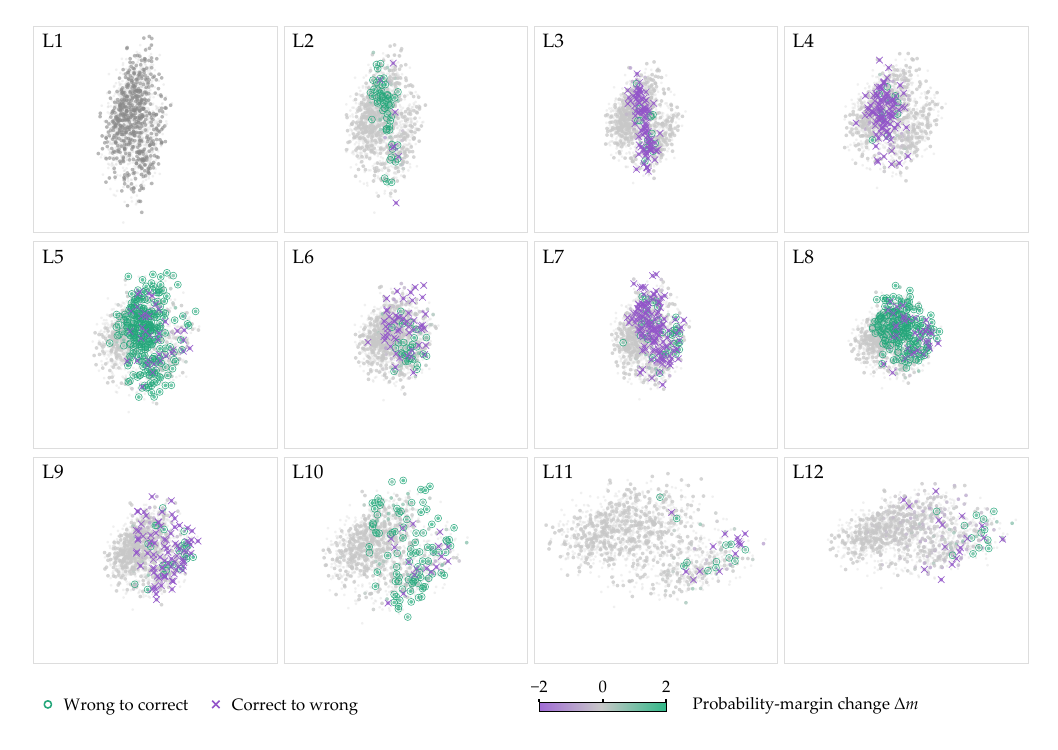}
\caption{\textbf{Our retrospective model on Is-this-a-good-customer.} All native layers in the fixed final-support PCA basis. Colors show adjacent probability-margin change under the frozen readout; rings and crosses mark corrections and deteriorations. The first layer is gray. Extraction and display conventions are given in \cref{app:extraction,app:visualization}.}\label{fig:customer-attnres}
\end{appendixfigure}\clearpage
\begin{appendixfigure}
\includegraphics[width=\linewidth,height=0.78\textheight,keepaspectratio]{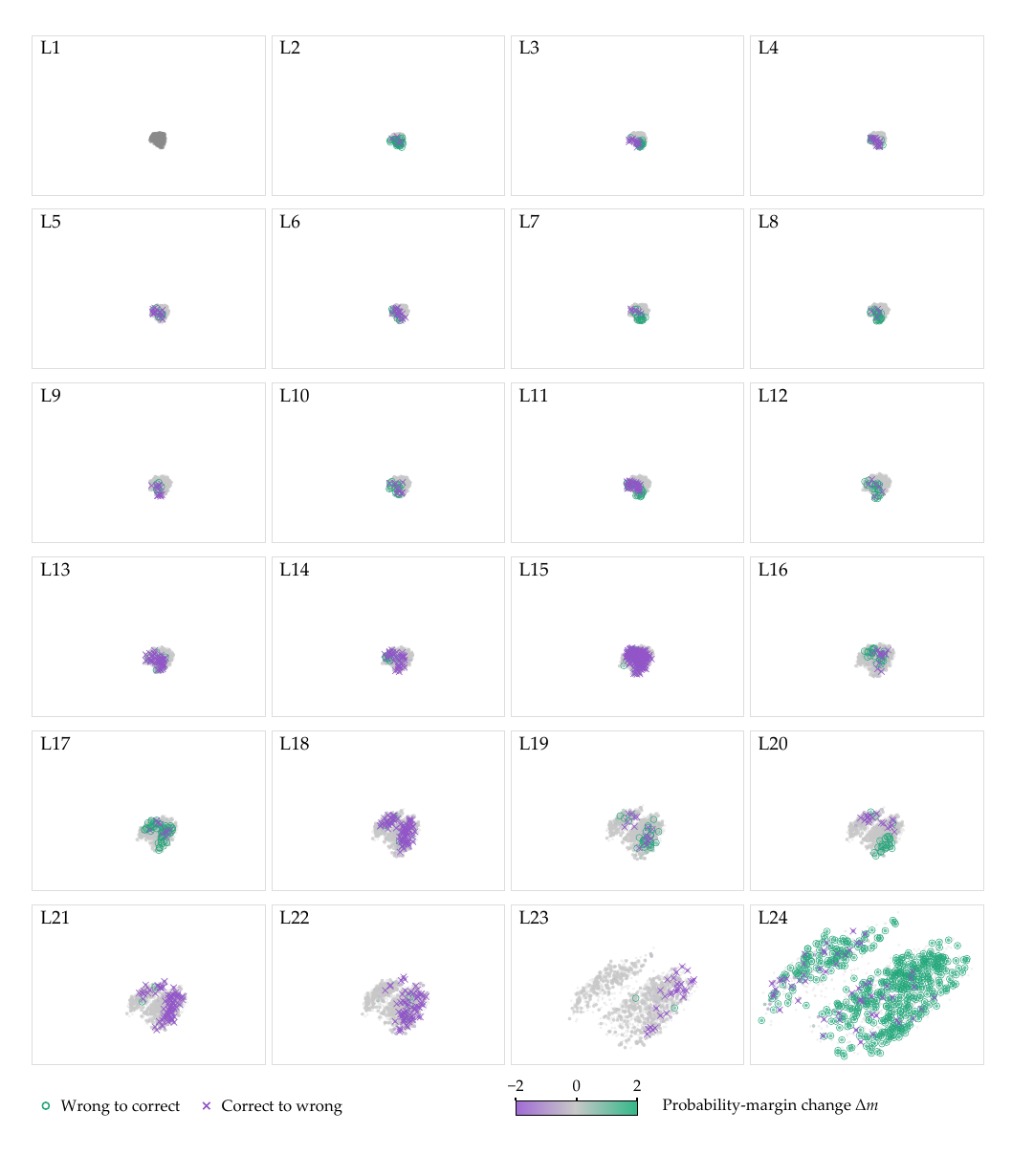}
\caption{\textbf{TabPFN-3 on Is-this-a-good-customer.} All native layers in the fixed final-support PCA basis. Colors show adjacent probability-margin change under the frozen readout; rings and crosses mark corrections and deteriorations. The first layer is gray. Extraction and display conventions are given in \cref{app:extraction,app:visualization}.}\label{fig:customer-tabpfn3}
\end{appendixfigure}\clearpage
\begin{appendixfigure}
\includegraphics[width=\linewidth,height=0.78\textheight,keepaspectratio]{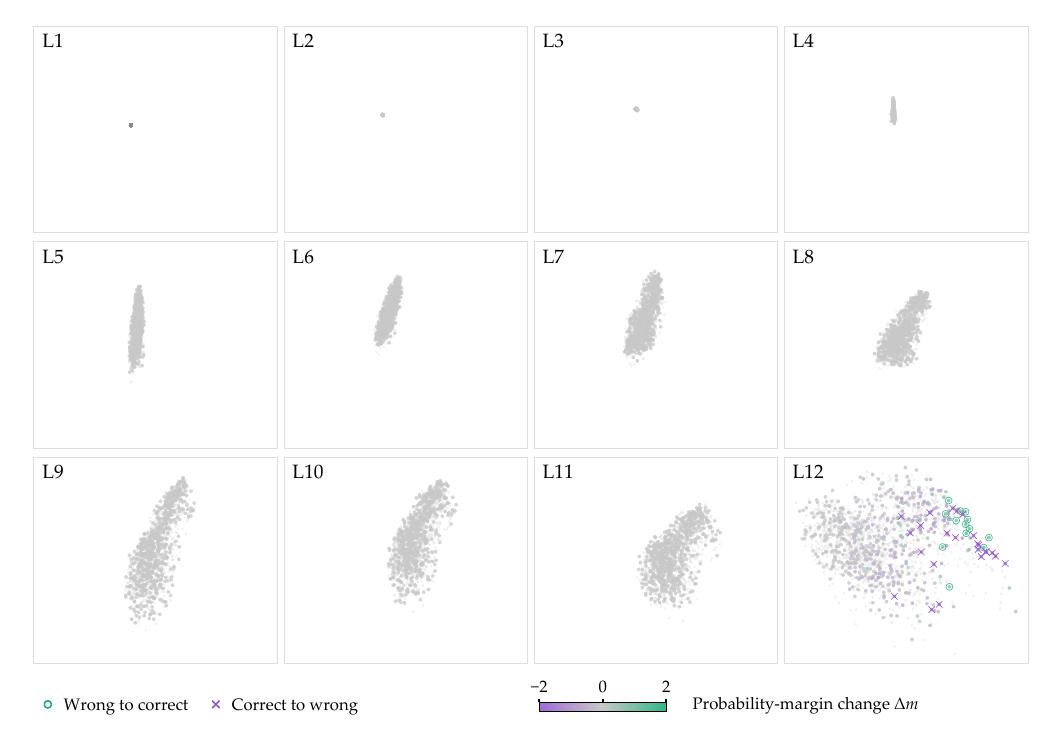}
\caption{\textbf{EXAONE-Tabular on Is-this-a-good-customer.} All native layers in the fixed final-support PCA basis. Colors show adjacent probability-margin change under the frozen readout; rings and crosses mark corrections and deteriorations. The first layer is gray. Extraction and display conventions are given in \cref{app:extraction,app:visualization}.}\label{fig:customer-exaone}
\end{appendixfigure}\clearpage

\clearpage
\section{Analysis of Gated Contextual Updates}
\label{app:gating-discussion}
\label{app:gating-interpretation}

This appendix extends the update-geometry analysis in
\cref{sec:selective}. We first test which aspects of the learned update
matter to a fixed predictor, then examine whether gates are more
interchangeable among nearby samples. The component comparison in
\cref{fig:component-ablation} is separate from these fixed-model interventions.
All results use native final prediction heads, unlike the frozen-readout
trajectories in \cref{app:protocol}.

\subsection{Direction and magnitude interventions}
\label{app:gating-geometry}
\paragraph{Protocol.}
We evaluate gated classification and regression models with 12 ICL layers,
width 512, and two-layer history groups. Each model uses weights averaged
over the final 100 updates of the second pretraining stage.
The study covers the first official split of all 51 TabArena datasets
(38 classification and 13 regression), using every support and query row,
one view, and FP32 inference without ensembling.
Query labels are used only for evaluation. Dataset means receive equal
weight; 95\% intervals use 20,000 paired dataset-bootstrap resamples.

\paragraph{Geometric controls.}
At each intervened state, let $o_{\ell i}$ and $g_{\ell i}$ be the
concatenated attention output and learned gate for query $i$.
Using the row-vector convention of \cref{eq:gate}, define
\begin{align}
 \bar g_\ell&=|Q|^{-1}\sum_{i\in Q}g_{\ell i},\nonumber\\
 u_{\ell i}&=(g_{\ell i}\odot o_{\ell i})W_{O,\ell}^{\top}
              +\mathbf b_{O,\ell}^{\top},\nonumber\\
 v_{\ell i}&=(\bar g_\ell\odot o_{\ell i})W_{O,\ell}^{\top}
              +\mathbf b_{O,\ell}^{\top}.
 \label{eq:gating-counterpart}
\end{align}
The counterpart $v_{\ell i}$ uses the channelwise mean gate of the current
unlabeled query batch. We compare replacing the update by $v_{\ell i}$
with two partial replacements:
\begin{equation}
 u_{\ell i}^{\mathrm{dir}}
   =v_{\ell i}\frac{\|u_{\ell i}\|_2}{\|v_{\ell i}\|_2},
 \qquad
 u_{\ell i}^{\mathrm{mag}}
   =u_{\ell i}\frac{\|v_{\ell i}\|_2}{\|u_{\ell i}\|_2}.
 \label{eq:gating-geometric-controls}
\end{equation}
The first changes direction while preserving the learned norm; the second
changes the norm while preserving direction. Both act on the projected
update, including its bias, rather than on the gate vector itself.
Only query updates are replaced, at all 12 layers. Support updates,
feed-forward computation, and normalization remain unchanged.
The pair $(u,v)$ is recomputed along each intervened trajectory.

\begin{figure}[!htb]
 \centering
 \includegraphics[width=\linewidth]{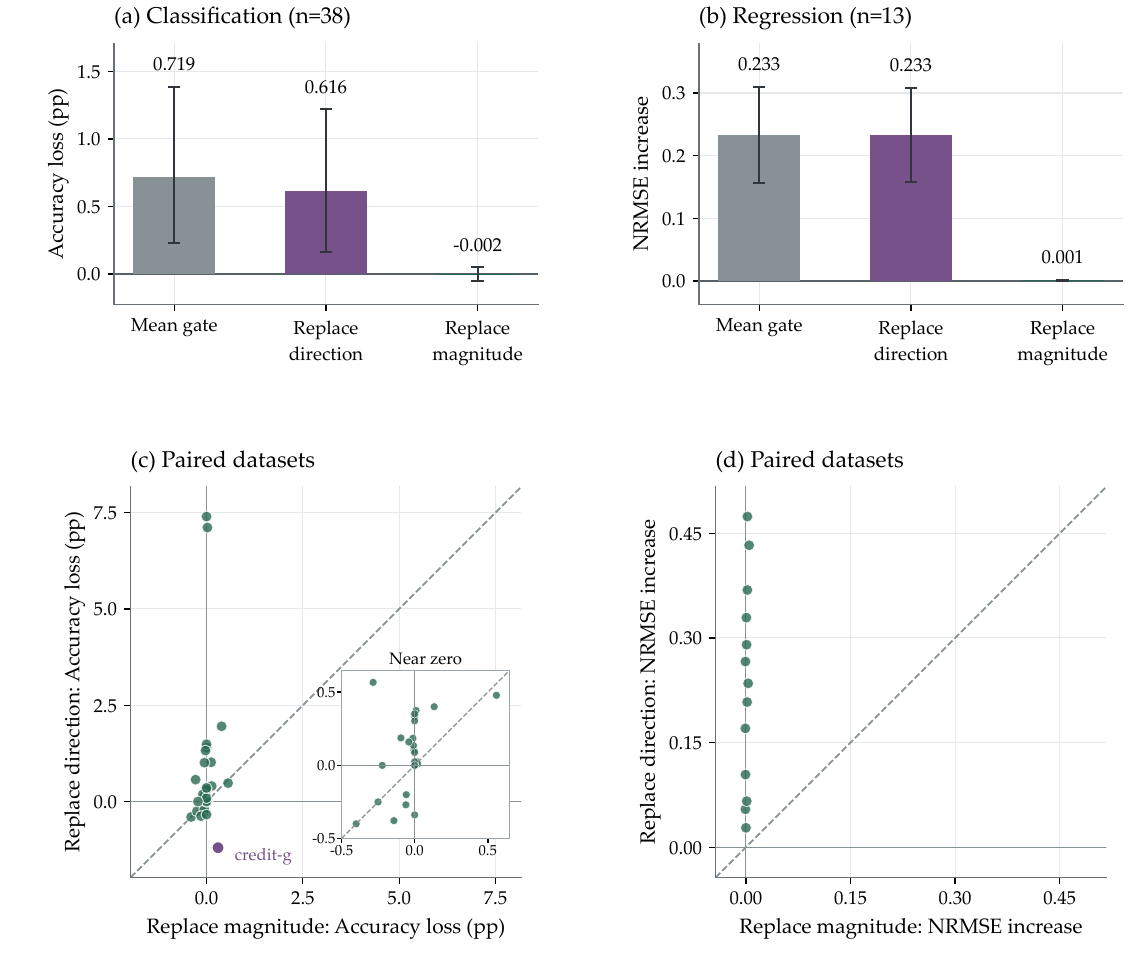}
 \caption{\textbf{Sensitivity to update direction and magnitude.} Top: mean degradation relative to learned gating, with 95\% dataset-bootstrap intervals. Bottom: dataset-level effects; points above the diagonal indicate greater harm from replacing direction. The classification inset enlarges the region near zero. Positive values mean lower accuracy or higher NRMSE. Replacement distances are not matched.}
 \label{fig:gated-update-geometry}
\end{figure}

\paragraph{Results.}
\Cref{fig:gated-update-geometry} expands \cref{fig:gated-incorporation}
with dataset-level effects. Mean-gate replacement reduces classification
accuracy by 0.719 percentage points. Replacing direction loses 0.616 points,
versus $-0.002$ for replacing magnitude. The paired contrast is
0.618 points (95\% interval: [0.173, 1.217]), but its median is 0.126:
23 datasets favor preserving direction, seven favor preserving magnitude,
and eight tie. On credit-g, direction replacement is less harmful by
1.497 points. The effect is therefore heterogeneous across classification
tasks.

For regression, NRMSE is query RMSE divided by the support-target standard
deviation. Mean-gate, direction, and magnitude replacement increase NRMSE
by 0.2326, 0.2328, and 0.0011, respectively.
The direction-minus-magnitude contrast is 0.2317
(95\% interval: [0.1579, 0.3070]), positive on all 13 datasets.
Under these controls, preserving the learned direction is less disruptive
than preserving only update size.

\paragraph{Interpretation.}
These results support query-conditioned shaping of contextual updates, particularly their direction. Together with the component comparison in \cref{fig:component-ablation}, they indicate that gating affects the geometry of the contextual update rather than only its overall scale.

\subsection{Gate exchange within sample neighborhoods}
\label{app:gating-locality}
We next examine whether gates can be exchanged with less disruption
between nearby samples. Normalized first-layer attention inputs are
standardized using support statistics. Eight MiniBatchKMeans clusters
are fitted to support rows, and queries are assigned without their labels.
Complete query gate vectors are cyclically exchanged within these groups.
The control uses random groups with exactly the same sizes.
Both conditions preserve the gate multiset, moved-query count, and
singleton count; support gates remain unchanged.
Groups are fixed across layers, and results are averaged over three
exchange seeds within each dataset.

In \cref{fig:gated-neighborhood-exchange}, neighborhood-constrained exchange
has a mean advantage of 0.441 accuracy points
(95\% interval: [0.057, 1.031]) and 0.0810 NRMSE
([0.0415, 0.1289]). Classification has 18 improvements, 14 reversals,
and six ties, with median advantage zero; all 13 regression datasets
favor neighborhood-constrained exchange.

Local exchanges also have smaller gate displacement on every dataset.
Mean gate RMS displacement is 0.0839 versus 0.0991 for classification
and 0.1163 versus 0.1322 for regression.
The smaller displacement helps explain why local exchanges are less
disruptive. Together, these observations are consistent with local
continuity of sample-conditioned updates, without establishing distinct
semantic roles for the neighborhoods.
\clearpage
\begin{appendixfigure}
  \includegraphics[width=\linewidth]{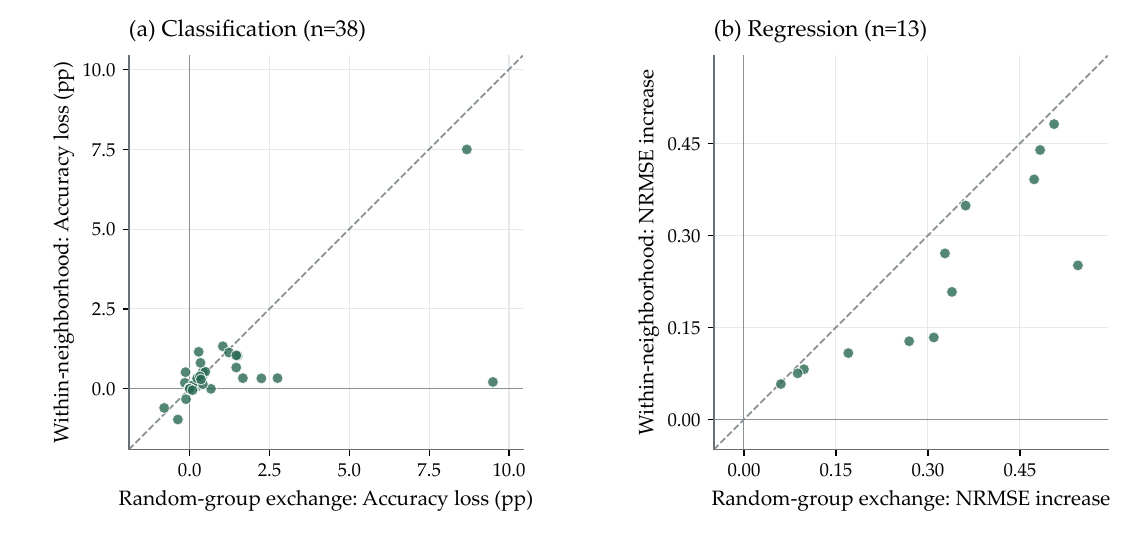}
  \caption{\textbf{Neighborhood-constrained versus random gate exchange.} Each point is a dataset, averaging three exchange seeds. Below the diagonal, exchange within support-derived neighborhoods is less harmful. Group sizes and moved-row counts are matched, within-neighborhood exchanges also induce smaller gate displacement.}
  \label{fig:gated-neighborhood-exchange}
\end{appendixfigure}

\clearpage
\section{Benchmark Protocols and Complete Rankings}
\label{app:benchmark-results}

This appendix provides the evaluation scope and complete rankings underlying
\cref{sec:retro,fig:benchmark-elo}. These results evaluate native predictors,
not the intermediate readouts in \cref{app:protocol} or the single-view
interventions in \cref{app:gating-discussion}.

\subsection{Evaluation scope}
\label{app:benchmark-scope}
TabArena includes 38 classification and 13 regression datasets, comprising
594 and 222 official splits, respectively.
TALENT includes 200 classification and 100 regression datasets.
RelArena includes 12 classification and nine regression tasks.
The comparison pools contain 37/38/39 methods for TabArena
(overall/classification/regression), 17 for each TALENT ranking, and
10 for each RelArena ranking.

\subsection{Ranking and uncertainty}
\label{app:benchmark-statistics}
All tables are ordered by decreasing Elo, with \Retro{} highlighted in bold.
Average Rank is lower-is-better. 
The benchmark-specific Elo procedures and comparison pools are retained,
so Elo values are not calibrated across benchmarks.

Elo+ and Elo$-$ are the distances from the point estimate to the upper
and lower interval endpoints. TabArena uses 200 dataset-bootstrap
replicates and RelArena 100 task-bootstrap replicates, with 2.5th and
97.5th percentile endpoints.
TALENT uses dataset-level aggregated scores, averaging evaluation seeds. Its Elo is averaged over 30 random
permutations of dataset order. The resulting interval reflects
order variability, not evaluation-seed uncertainty or a bootstrap
confidence interval.

For TabArena, Average Rank is computed from ranks within each split,
averaged first within a dataset and then equally across datasets.
TALENT and RelArena retain their benchmark-exported average ranks.
Overall Elo is fitted to the pooled classification and regression results.

\paragraph{Component comparison.}
\label{app:component-comparison}
\Cref{fig:component-ablation} compares both mechanisms, Attention Residuals
only, Gated Attention only, and neither on TabArena classification, with
Elo point estimates 1115.6, 1093.1, 969.0, and 822.3, respectively.
These values belong to the four-configuration comparison, not the full
benchmark candidate pool, and should not be compared directly with
\cref{fig:benchmark-elo}. Whiskers reproduce the reported 95\% intervals.

\clearpage
\subsection{TabArena}
\label{app:rankings-tabarena}
\Cref{tab:elo-tabarena-pairs,tab:elo-tabarena-classification,tab:elo-tabarena-regression}
report the complete overall, classification, and regression comparisons
for the TabArena panels of \cref{fig:benchmark-elo}.
\begin{table}[!htbp]
  \centering
  \setlength{\tabcolsep}{11.5pt}
  \renewcommand{\arraystretch}{1.03}
  \caption{\textbf{TabArena: overall ranking.} All 37 selected methods on 51 datasets.}
  \label{tab:elo-tabarena-pairs}
  \vspace{5pt}
  \begin{tabular}{lcccc}
  \toprule
  Method & Elo $\uparrow$ & Elo+ & Elo$-$ & Average Rank $\downarrow$ \\
  \midrule
  TabFM (default) & 1882.9 & 109.9 & 99.5 & 4.31 \\
  EXAONE-Tabular (default) & 1849.7 & 72.8 & 49.7 & 4.89 \\
  \textbf{\Retro{}} & \textbf{1741.6} & \textbf{84.3} & \textbf{72.5} & \textbf{7.26} \\
  TabPFN-3 (default) & 1728.9 & 76.1 & 55.1 & 7.58 \\
  Xiaomi-TabLDM (default) & 1673.7 & 74.2 & 66.0 & 9.11 \\
  TabPFN-2.6 (default) & 1669.5 & 62.7 & 44.9 & 9.24 \\
  TabICLv2 (default) & 1650.9 & 67.6 & 57.6 & 9.80 \\
  TabDPT-1.3 (default) & 1601.4 & 60.6 & 40.1 & 11.39 \\
  RealMLP (tuned + ensembled) & 1558.3 & 46.1 & 44.7 & 12.88 \\
  TabDPT-Turbo (default) & 1502.2 & 59.1 & 50.6 & 14.93 \\
  TabM (tuned + ensembled) & 1501.3 & 49.8 & 45.7 & 14.97 \\
  LightGBM (tuned + ensembled) & 1483.0 & 32.0 & 33.5 & 15.66 \\
  RealMLP (tuned) & 1470.7 & 46.5 & 50.4 & 16.13 \\
  CatBoost (tuned + ensembled) & 1467.6 & 33.1 & 39.8 & 16.25 \\
  CatBoost (tuned) & 1452.1 & 36.5 & 35.0 & 16.85 \\
  TabM (tuned) & 1437.1 & 49.6 & 43.0 & 17.43 \\
  ModernNCA (tuned + ensembled) & 1437.1 & 67.8 & 52.1 & 17.43 \\
  LightGBM (tuned) & 1427.2 & 31.4 & 26.7 & 17.82 \\
  XGBoost (tuned + ensembled) & 1417.9 & 32.0 & 31.9 & 18.18 \\
  CatBoost (default) & 1410.3 & 47.6 & 43.1 & 18.48 \\
  ModernNCA (tuned) & 1394.9 & 41.1 & 40.3 & 19.08 \\
  XGBoost (tuned) & 1387.8 & 30.3 & 30.9 & 19.36 \\
  TabSwift (default) & 1387.8 & 66.3 & 53.8 & 19.36 \\
  TabM (default) & 1328.5 & 48.2 & 47.5 & 21.65 \\
  ModernNCA (default) & 1279.4 & 39.2 & 40.3 & 23.50 \\
  ExtraTrees (tuned + ensembled) & 1254.1 & 53.4 & 55.9 & 24.42 \\
  RealMLP (default) & 1253.7 & 36.1 & 45.8 & 24.43 \\
  XGBoost (default) & 1234.3 & 37.6 & 41.9 & 25.12 \\
  ExtraTrees (tuned) & 1218.7 & 58.2 & 68.1 & 25.66 \\
  RandomForest (tuned + ensembled) & 1207.4 & 61.8 & 58.3 & 26.04 \\
  LightGBM (default) & 1202.6 & 43.4 & 36.9 & 26.20 \\
  RandomForest (tuned) & 1163.2 & 62.9 & 69.3 & 27.48 \\
  Linear (tuned + ensembled) & 1032.9 & 65.0 & 106.2 & 31.10 \\
  ExtraTrees (default) & 1019.3 & 56.9 & 86.3 & 31.42 \\
  RandomForest (default) & 1000.0 & 55.0 & 66.1 & 31.86 \\
  Linear (tuned) & 996.6 & 70.0 & 126.9 & 31.93 \\
  Linear (default) & 897.1 & 76.7 & 148.5 & 33.81 \\
  \bottomrule
  \end{tabular}
\end{table}

\clearpage
\begin{table}[!htbp]
  \centering
  \setlength{\tabcolsep}{11.5pt}
  \renewcommand{\arraystretch}{1.03}
  \caption{\textbf{TabArena: classification ranking.} All 38 selected methods on 38 datasets.}
  \label{tab:elo-tabarena-classification}
  \vspace{5pt}
  \begin{tabular}{lcccc}
  \toprule
  Method & Elo $\uparrow$ & Elo+ & Elo$-$ & Average Rank $\downarrow$ \\
  \midrule
  TabFM (default) & 1833.3 & 118.3 & 107.4 & 4.76 \\
  EXAONE-Tabular (default) & 1821.1 & 82.0 & 56.0 & 5.00 \\
  TabPFN-3 (default) & 1690.7 & 74.8 & 68.0 & 8.12 \\
  \textbf{\Retro{}} & \textbf{1680.4} & \textbf{79.1} & \textbf{70.8} & \textbf{8.42} \\
  TabPFN-2.6 (default) & 1638.4 & 56.6 & 50.2 & 9.70 \\
  Xiaomi-TabLDM (default) & 1631.5 & 70.7 & 60.1 & 9.93 \\
  TabICLv2 (default) & 1627.6 & 78.4 & 68.7 & 10.05 \\
  TabDPT-1.3 (default) & 1566.9 & 83.6 & 61.4 & 12.16 \\
  RealMLP (tuned + ensembled) & 1516.8 & 50.0 & 41.7 & 14.05 \\
  TabM (tuned + ensembled) & 1495.9 & 57.1 & 45.0 & 14.88 \\
  LightGBM (tuned + ensembled) & 1463.0 & 51.0 & 35.2 & 16.21 \\
  TabDPT-Turbo (default) & 1454.3 & 72.5 & 60.1 & 16.57 \\
  TabICL (default) & 1447.4 & 49.0 & 56.2 & 16.85 \\
  CatBoost (tuned + ensembled) & 1442.0 & 50.1 & 42.7 & 17.07 \\
  TabM (tuned) & 1437.8 & 55.9 & 43.0 & 17.25 \\
  RealMLP (tuned) & 1434.3 & 56.6 & 54.2 & 17.40 \\
  CatBoost (tuned) & 1430.9 & 49.0 & 44.7 & 17.54 \\
  LightGBM (tuned) & 1413.3 & 41.1 & 34.8 & 18.27 \\
  XGBoost (tuned + ensembled) & 1407.7 & 42.0 & 47.0 & 18.51 \\
  CatBoost (default) & 1407.2 & 48.3 & 47.6 & 18.53 \\
  ModernNCA (tuned + ensembled) & 1391.6 & 71.9 & 65.9 & 19.18 \\
  ModernNCA (tuned) & 1385.3 & 51.3 & 35.9 & 19.44 \\
  XGBoost (tuned) & 1376.6 & 38.1 & 46.0 & 19.81 \\
  TabSwift (default) & 1366.2 & 64.6 & 61.1 & 20.25 \\
  TabM (default) & 1331.4 & 53.4 & 48.0 & 21.69 \\
  ModernNCA (default) & 1265.9 & 45.2 & 48.3 & 24.34 \\
  RealMLP (default) & 1251.2 & 39.7 & 45.2 & 24.92 \\
  XGBoost (default) & 1242.9 & 43.7 & 50.1 & 25.24 \\
  ExtraTrees (tuned + ensembled) & 1241.7 & 63.5 & 62.3 & 25.28 \\
  RandomForest (tuned + ensembled) & 1210.6 & 77.4 & 77.2 & 26.45 \\
  ExtraTrees (tuned) & 1207.0 & 67.6 & 73.2 & 26.58 \\
  LightGBM (default) & 1192.7 & 57.8 & 55.7 & 27.10 \\
  RandomForest (tuned) & 1166.1 & 76.0 & 80.3 & 28.03 \\
  Linear (tuned + ensembled) & 1088.1 & 91.1 & 98.4 & 30.51 \\
  Linear (tuned) & 1053.4 & 91.8 & 106.5 & 31.48 \\
  RandomForest (default) & 1000.0 & 78.3 & 78.3 & 32.81 \\
  ExtraTrees (default) & 997.5 & 90.5 & 87.7 & 32.87 \\
  Linear (default) & 956.3 & 97.9 & 123.9 & 33.76 \\
  \bottomrule
  \end{tabular}
\end{table}

\clearpage
\begin{table}[!htbp]
  \centering
  \setlength{\tabcolsep}{11pt}
  \renewcommand{\arraystretch}{1.03}
  \caption{\textbf{TabArena: regression ranking.} All 39 selected methods on 13 datasets.}
  \label{tab:elo-tabarena-regression}
  \vspace{5pt}
  \begin{tabular}{lcccc}
  \toprule
  Method & Elo $\uparrow$ & Elo+ & Elo$-$ & Average Rank $\downarrow$ \\
  \midrule
  TabFM (default) & 2165.6 & 254.1 & 73.4 & 3.56 \\
  \textbf{\Retro{}} & \textbf{2082.1} & \textbf{332.6} & \textbf{117.5} & \textbf{4.83} \\
  EXAONE-Tabular (default) & 2043.9 & 251.0 & 83.9 & 5.52 \\
  TabPFN-3 (default) & 1966.4 & 320.5 & 122.8 & 7.15 \\
  Xiaomi-TabLDM (default) & 1928.5 & 348.6 & 176.1 & 8.05 \\
  Nori-30M (default) & 1895.0 & 254.2 & 76.1 & 8.90 \\
  TabPFN-2.6 (default) & 1866.7 & 252.9 & 51.6 & 9.66 \\
  TabICLv2 (default) & 1835.8 & 292.9 & 124.4 & 10.53 \\
  Nori (default) & 1825.1 & 242.2 & 65.6 & 10.84 \\
  TabDPT-1.3 (default) & 1817.0 & 299.2 & 111.4 & 11.07 \\
  RealMLP (tuned + ensembled) & 1786.9 & 204.5 & 47.5 & 11.97 \\
  TabDPT-Turbo (default) & 1762.6 & 251.9 & 83.4 & 12.72 \\
  ModernNCA (tuned + ensembled) & 1684.5 & 206.6 & 99.9 & 15.24 \\
  RealMLP (tuned) & 1678.3 & 189.1 & 78.3 & 15.45 \\
  CatBoost (tuned + ensembled) & 1628.3 & 212.6 & 76.2 & 17.13 \\
  LightGBM (tuned + ensembled) & 1623.3 & 202.9 & 90.5 & 17.30 \\
  CatBoost (tuned) & 1597.8 & 219.6 & 88.8 & 18.17 \\
  TabM (tuned + ensembled) & 1593.0 & 240.3 & 96.3 & 18.34 \\
  LightGBM (tuned) & 1546.3 & 227.3 & 87.7 & 19.94 \\
  TabSwift (default) & 1536.7 & 212.2 & 140.0 & 20.27 \\
  XGBoost (tuned + ensembled) & 1523.8 & 195.5 & 77.9 & 20.71 \\
  ModernNCA (tuned) & 1509.2 & 199.9 & 105.4 & 21.21 \\
  TabM (tuned) & 1505.1 & 248.6 & 102.7 & 21.34 \\
  XGBoost (tuned) & 1497.4 & 217.6 & 87.9 & 21.61 \\
  CatBoost (default) & 1491.9 & 167.0 & 126.8 & 21.79 \\
  ModernNCA (default) & 1393.3 & 192.3 & 94.7 & 25.01 \\
  TabM (default) & 1384.0 & 222.6 & 125.6 & 25.30 \\
  ExtraTrees (tuned + ensembled) & 1354.7 & 168.4 & 107.1 & 26.20 \\
  RealMLP (default) & 1325.6 & 172.1 & 125.8 & 27.05 \\
  ExtraTrees (tuned) & 1316.4 & 169.4 & 113.4 & 27.31 \\
  LightGBM (default) & 1288.8 & 168.9 & 90.4 & 28.08 \\
  XGBoost (default) & 1259.9 & 191.5 & 124.5 & 28.84 \\
  RandomForest (tuned + ensembled) & 1246.1 & 118.0 & 116.6 & 29.19 \\
  RandomForest (tuned) & 1199.1 & 119.6 & 129.5 & 30.31 \\
  ExtraTrees (default) & 1121.8 & 133.6 & 157.9 & 31.91 \\
  RandomForest (default) & 1000.0 & 107.5 & 137.5 & 33.82 \\
  Linear (tuned + ensembled) & 491.6 & 142.7 & 2134.7 & 37.27 \\
  Linear (tuned) & 389.4 & 185.5 & 2245.4 & 37.71 \\
  Linear (default) & 121.4 & 220.6 & 2245.2 & 38.69 \\
  \bottomrule
  \end{tabular}
\end{table}

\clearpage

\subsection{TALENT}
\label{app:rankings-talent}
\Cref{tab:elo-talent-pairs,tab:elo-talent-classification,tab:elo-talent-regression}
report the corresponding TALENT comparisons. Their interval columns use
the dataset-order variability described in \cref{app:benchmark-statistics}.
\begin{table}[!htbp]
  \centering
  \setlength{\tabcolsep}{11pt}
  \renewcommand{\arraystretch}{1.03}
  \caption{\textbf{TALENT: overall ranking.} All 17 selected methods on 300 datasets.}
  \label{tab:elo-talent-pairs}
  \vspace{5pt}
  \begin{tabular}{lcccc}
  \toprule
  Method & Elo $\uparrow$ & Elo+ & Elo$-$ & Average Rank $\downarrow$ \\
  \midrule
  TabFM & 1856.0 & 338.5 & 339.8 & 3.99 \\
  \textbf{\Retro{}} & \textbf{1755.4} & \textbf{280.9} & \textbf{443.1} & \textbf{5.41} \\
  EXAONE-Tabular & 1745.2 & 254.4 & 239.3 & 5.22 \\
  Xiaomi-TabLDM & 1717.2 & 223.6 & 290.2 & 6.05 \\
  TabPFN-3 & 1697.7 & 357.0 & 347.5 & 5.68 \\
  TabICLv2 & 1679.3 & 218.9 & 384.4 & 6.22 \\
  TabPFN-2.5 & 1614.0 & 251.5 & 334.8 & 7.01 \\
  TabPFN-2.6 & 1566.5 & 277.9 & 329.9 & 6.91 \\
  TabSwift & 1432.0 & 276.8 & 295.7 & 9.80 \\
  TabM (tuned) & 1394.0 & 415.9 & 282.2 & 11.08 \\
  ModernNCA (tuned) & 1385.8 & 342.0 & 334.2 & 10.85 \\
  CatBoost (tuned) & 1383.4 & 287.3 & 301.1 & 11.18 \\
  LightGBM (tuned) & 1364.5 & 341.4 & 250.8 & 11.74 \\
  RealMLP (tuned) & 1308.6 & 318.5 & 230.9 & 11.64 \\
  XGBoost (tuned) & 1289.7 & 324.0 & 315.4 & 12.32 \\
  RandomForest (tuned) & 1239.0 & 546.8 & 365.2 & 13.80 \\
  MLP (tuned) & 1071.8 & 226.8 & 239.0 & 14.10 \\
  \bottomrule
  \end{tabular}
\end{table}

\begin{table}[!htbp]
  \centering
  \setlength{\tabcolsep}{11pt}
  \renewcommand{\arraystretch}{1.03}
  \caption{\textbf{TALENT: classification ranking.} All 17 selected methods on 200 datasets.}
  \label{tab:elo-talent-classification}
  \vspace{5pt}
  \begin{tabular}{lcccc}
  \toprule
  Method & Elo $\uparrow$ & Elo+ & Elo$-$ & Average Rank $\downarrow$ \\
  \midrule
  TabFM & 1829.8 & 306.9 & 408.4 & 4.11 \\
  \textbf{\Retro{}} & \textbf{1719.4} & \textbf{312.6} & \textbf{299.9} & \textbf{5.89} \\
  TabICLv2 & 1715.2 & 316.5 & 362.1 & 5.93 \\
  EXAONE-Tabular & 1708.3 & 238.5 & 255.4 & 5.53 \\
  TabPFN-3 & 1676.8 & 267.1 & 228.3 & 5.80 \\
  Xiaomi-TabLDM & 1670.8 & 304.8 & 295.4 & 6.37 \\
  TabPFN-2.5 & 1633.4 & 308.0 & 290.4 & 6.72 \\
  TabPFN-2.6 & 1571.9 & 246.6 & 340.5 & 7.07 \\
  TabM (tuned) & 1464.3 & 401.5 & 393.3 & 10.51 \\
  TabSwift & 1417.8 & 272.4 & 203.6 & 9.60 \\
  CatBoost (tuned) & 1408.6 & 325.2 & 279.8 & 11.37 \\
  LightGBM (tuned) & 1363.1 & 362.6 & 375.2 & 11.69 \\
  ModernNCA (tuned) & 1321.8 & 411.2 & 252.9 & 10.82 \\
  RealMLP (tuned) & 1313.4 & 354.4 & 238.9 & 11.64 \\
  XGBoost (tuned) & 1282.7 & 307.1 & 206.8 & 12.18 \\
  RandomForest (tuned) & 1232.4 & 441.7 & 340.8 & 13.87 \\
  MLP (tuned) & 1170.5 & 368.5 & 307.7 & 13.91 \\
  \bottomrule
  \end{tabular}
\end{table}

\clearpage
\input{tab/benchmark_talent_regression}
\clearpage

\subsection{RelArena}
\label{app:rankings-relarena}
\Cref{tab:elo-relarena-pairs,tab:elo-relarena-classification,tab:elo-relarena-regression}
report the RelArena comparisons. Overall results pool all 21 tasks, while
the two task-specific tables retain their respective subsets.
\begin{table}[!htbp]
  \centering
  \setlength{\tabcolsep}{11pt}
  \renewcommand{\arraystretch}{1.03}
  \caption{\textbf{RelArena: overall ranking.} All 10 selected methods on 21 tasks.}
  \label{tab:elo-relarena-pairs}
  \vspace{5pt}
  \begin{tabular}{lcccc}
  \toprule
  Method & Elo $\uparrow$ & Elo+ & Elo$-$ & Average Rank $\downarrow$ \\
  \midrule
  \textbf{\Retro{}} & \textbf{1797.0} & \textbf{78.4} & \textbf{60.4} & \textbf{2.90} \\
  KurveRSC & 1781.2 & 92.8 & 93.6 & 3.05 \\
  TabPFN-rel-local & 1721.8 & 116.2 & 80.2 & 3.62 \\
  GraphSAGE & 1666.4 & 90.8 & 78.6 & 4.19 \\
  RelGT & 1575.2 & 126.8 & 103.3 & 5.17 \\
  RDBLearn & 1555.1 & 89.6 & 93.1 & 5.38 \\
  RelGNN-ES & 1521.4 & 90.6 & 81.6 & 5.74 \\
  LightGBM & 1356.2 & 90.2 & 120.8 & 7.33 \\
  Constant (Per-entity) & 1260.3 & 130.4 & 150.4 & 8.10 \\
  Constant (Global) & 1000.0 & 70.9 & 130.1 & 9.52 \\
  \bottomrule
  \end{tabular}
\end{table}

\begin{table}[!htbp]
  \centering
  \setlength{\tabcolsep}{11pt}
  \renewcommand{\arraystretch}{1.03}
  \caption{\textbf{RelArena: classification ranking.} All 10 selected methods on 12 tasks.}
  \label{tab:elo-relarena-classification}
  \vspace{5pt}
  \begin{tabular}{lcccc}
  \toprule
  Method & Elo $\uparrow$ & Elo+ & Elo$-$ & Average Rank $\downarrow$ \\
  \midrule
  \textbf{\Retro{}} & \textbf{2185.9} & \textbf{266.1} & \textbf{45.3} & \textbf{3.08} \\
  KurveRSC & 2151.6 & 340.0 & 111.5 & 3.42 \\
  TabPFN-rel-local & 2135.2 & 315.9 & 122.3 & 3.58 \\
  GraphSAGE & 2103.2 & 358.6 & 135.7 & 3.92 \\
  RelGT & 1981.0 & 352.6 & 124.4 & 5.25 \\
  RelGNN-ES & 1958.1 & 292.1 & 106.1 & 5.50 \\
  RDBLearn & 1958.1 & 292.1 & 89.5 & 5.50 \\
  LightGBM & 1770.2 & 250.6 & 106.4 & 7.33 \\
  Constant (Per-entity) & 1754.8 & 331.8 & 185.6 & 7.46 \\
  Constant (Global) & 1000.0 & 186.4 & 1952.8 & 9.96 \\
  \bottomrule
  \end{tabular}
\end{table}

\begin{table}[!htbp]
  \centering
  \setlength{\tabcolsep}{11pt}
  \renewcommand{\arraystretch}{1.03}
  \caption{\textbf{RelArena: regression ranking.} All 10 selected methods on 9 tasks.}
  \label{tab:elo-relarena-regression}
  \vspace{5pt}
  \begin{tabular}{lcccc}
  \toprule
  Method & Elo $\uparrow$ & Elo+ & Elo$-$ & Average Rank $\downarrow$ \\
  \midrule
  KurveRSC & 1736.4 & 321.6 & 143.4 & 2.56 \\
  \textbf{\Retro{}} & \textbf{1722.3} & \textbf{302.3} & \textbf{130.4} & \textbf{2.67} \\
  TabPFN-rel-local & 1609.0 & 247.2 & 169.5 & 3.67 \\
  GraphSAGE & 1519.0 & 103.4 & 94.2 & 4.56 \\
  RelGT & 1469.8 & 208.6 & 185.5 & 5.06 \\
  RDBLearn & 1453.4 & 158.5 & 188.5 & 5.22 \\
  RelGNN-ES & 1369.7 & 162.9 & 197.9 & 6.06 \\
  LightGBM & 1228.9 & 103.2 & 211.6 & 7.33 \\
  Constant (Per-entity) & 1000.0 & 184.3 & 353.1 & 8.94 \\
  Constant (Global) & 1000.0 & 63.7 & 147.0 & 8.94 \\
  \bottomrule
  \end{tabular}
\end{table}

\end{document}